\documentclass[10pt,twocolumn,letterpaper]{article}

\usepackage{iccv}
\usepackage{times}
\usepackage{epsfig}
\usepackage{graphicx}
\usepackage{amsmath}
\usepackage{amssymb}

\usepackage{booktabs}
\usepackage{bbding}
\usepackage{multirow}
\usepackage{caption}
\usepackage{subcaption}
\usepackage[accsupp]{axessibility}

\usepackage[pagebackref=true,breaklinks=true,colorlinks,bookmarks=false]{hyperref}

\usepackage[capitalize]{cleveref}
\crefname{section}{Sec.}{Secs.}
\Crefname{section}{Section}{Sections}
\Crefname{table}{Table}{Tables}
\crefname{table}{Tab.}{Tabs.}

 \iccvfinalcopy 

\def\iccvPaperID{5441} 

\ificcvfinal\fi

\begin{document}
	
	\title{On the Effectiveness of Spectral Discriminators for\\ Perceptual Quality Improvement}
	
	\author{
		Xin Luo~~~~Yunan Zhu~~~~Shunxin Xu~~~~Dong Liu\\
		University of Science and Technology of China, Hefei, China\\
		{\tt\small \{xinluo, zhuyn, sxu\}@mail.ustc.edu.cn, \tt\small dongeliu@ustc.edu.cn}
		\\\small\textbf{\url{https://github.com/Luciennnnnnn/DualFormer}}
}

\maketitle
\ificcvfinal\thispagestyle{empty}\fi

\captionsetup[figure]{font=small}
\captionsetup[table]{font=small}
\captionsetup[sub]{font=small}

\begin{abstract}
	\let\thefootnote\relax\footnote{This work was supported by the Natural Science Foundation of China under Grants 62022075 and 62036005, and by the Fundamental Research Funds for the Central Universities under Grant WK3490000006. This work was also supported by the advanced computing resources provided by the Supercomputing Center of USTC.~\emph{(Corresponding author: Dong Liu.)}}

	Several recent studies advocate the use of spectral discriminators, which evaluate the Fourier spectra of images for generative modeling. However, the effectiveness of the spectral discriminators is not well interpreted yet. We tackle this issue by examining the spectral discriminators in the context of perceptual image super-resolution~(i.e., GAN-based SR), as SR image quality is susceptible to spectral changes. Our analyses reveal that the spectral discriminator indeed performs better than the ordinary~(a.k.a. spatial) discriminator in identifying the differences in the \emph{high-frequency} range; however, the spatial discriminator holds an advantage in the \emph{low-frequency} range. Thus, we suggest that the spectral and spatial discriminators shall be used simultaneously. Moreover, we improve the spectral discriminators by first calculating the patch-wise Fourier spectrum and then aggregating the spectra by Transformer. We verify the effectiveness of the proposed method twofold. On the one hand, thanks to the additional spectral discriminator, our obtained SR images have their spectra better aligned to those of the real images, which leads to a better PD tradeoff. On the other hand, our ensembled discriminator predicts the perceptual quality more accurately, as evidenced in the no-reference image quality assessment task. \vspace{-1em}
\end{abstract}

\section{Introduction}
\label{sec:intro}

\renewcommand{\thefootnote}{\arabic{footnote}}

\begin{figure}[t]
	\setlength{\abovecaptionskip}{3pt}
	\centering
	\begin{subfigure}[b]{0.95\linewidth}
		\setlength{\abovecaptionskip}{3pt}
		\centering
		\includegraphics[width=\textwidth]{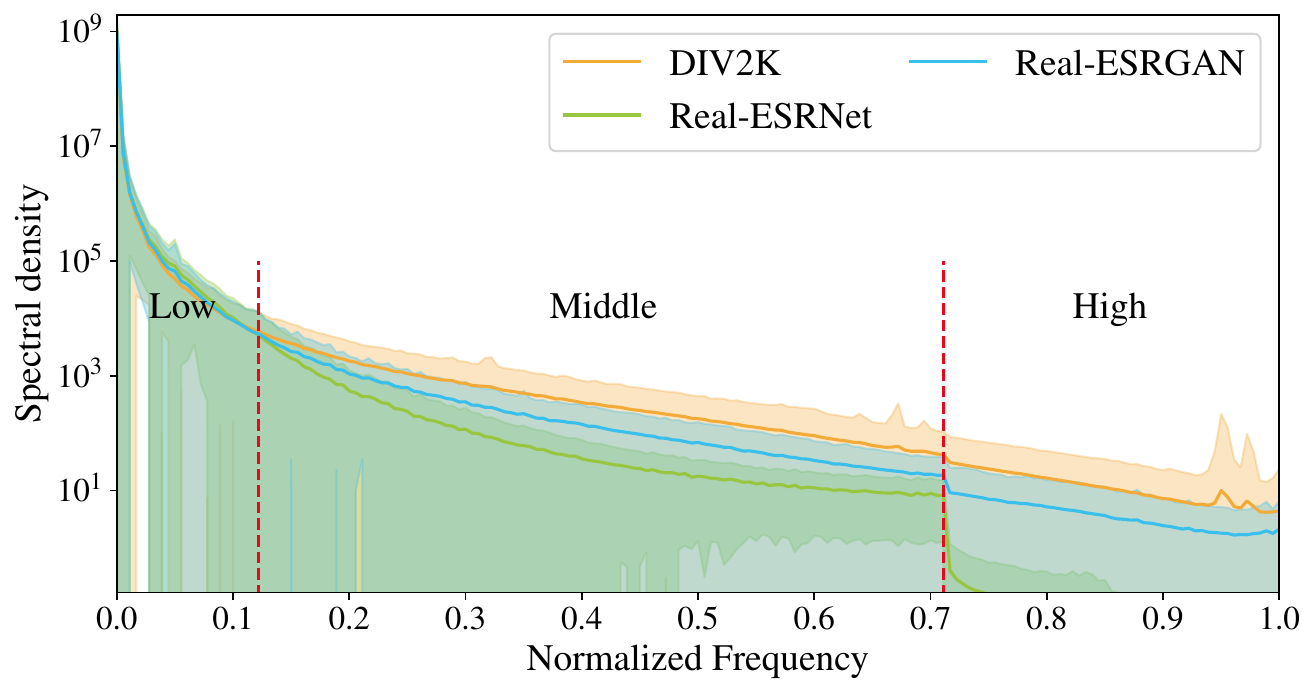}
		\caption{Spectra Statistics}
		\label{fig:fig1:a}
	\end{subfigure}
	
	\begin{subfigure}[b]{0.3\linewidth}
		\setlength{\abovecaptionskip}{3pt}
		\centering
		\includegraphics[width=\linewidth]{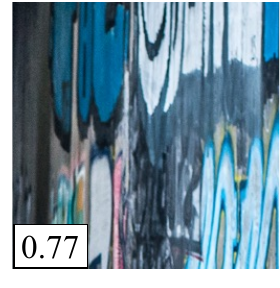}
		\caption{Ground Truth}
		\label{fig:fig1:b}
	\end{subfigure}
	\begin{subfigure}[b]{0.3\linewidth}
		\setlength{\abovecaptionskip}{3pt}
		\centering
		\includegraphics[width=\linewidth]{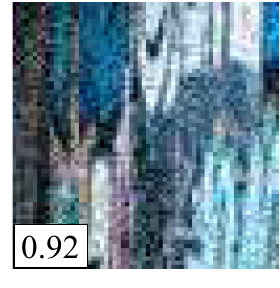}
		\caption{Low-Quality}
		\label{fig:fig1:c}
	\end{subfigure}
	\begin{subfigure}[b]{0.3\linewidth}
		\setlength{\abovecaptionskip}{3pt}
		\centering
		\includegraphics[width=\linewidth]{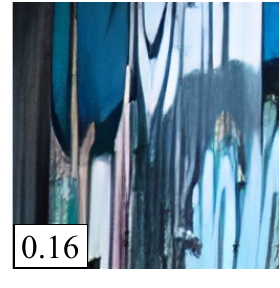}
		\caption{Super-Resolved}
		\label{fig:fig1:d}
	\end{subfigure}
	
	\caption{Denote the reduced spectrum $\tilde{S}\left(r\right)$ as the azimuthal average over the spectrum in normalized polar coordinates $r \in\left[0,1\right]$. (a) shows spectra statistics $\mathbb{E}[\tilde{S}\left(r\right)]$ of Real-ESRNet~\cite{wang2021real} and Real-ESRGAN~\cite{wang2021real} on the DIV2K~\cite{agustsson2017ntire} dataset. It illustrates that Real-ESRGAN improves perceptual quality by producing more high-frequency information. (b), (c) and (d) present the average scores of Real-ESRGAN's discriminator \wrt three types of images. Given second-order degradation model~\cite{wang2021real} to generate low-resolution input, (c) are bicubic upsampled from low-resolution input, and (d) are generated by the generator, which takes low-resolution images as input. The extremely high score on (c) indicates that spatial discriminators may excessively favor high-frequency information, even if it is noise or artifacts.}
	\label{fig:fig1}
	\vspace{-1.5em}
\end{figure}

Generative Adversarial Networks~(GANs)~\cite{goodfellow2014generative} have gained widespread adoption in low-level vision tasks, \eg, image super-resolution~(SR)~\cite{ledig2017photo}, which improves the perceptual quality of reconstructed images considerably. From the perspective of PD tradeoff~\cite{blau2018perception}, GANs improve perceptual quality by measuring the difference between image distributions~(perception) rather than signal differences~(distortion). In addition, Zhu~\etal~\cite{zhu2021recycling} found that trained discriminators can perform no-reference image quality assessment~(NR-IQA). Therefore, the key to improving the perceptual quality in GAN-based SR is to enhance the discriminator's ability to measure the differences in image distributions, which can predict image quality more accurately.

Recently, many studies indicate that images produced by GAN have difficulty matching the spectral distribution of real data, especially in the high-frequency range~\cite{durall2020watch,frank2020leveraging,khayatkhoei2022spatial,dzanic2020fourier,wang2020cnn,chandrasegaran2021closer,schwarz2021frequency}. Spectral discriminators~\cite{chen2021ssd,jung2021spectral,schwarz2021frequency} that utilize the Fourier spectrum as input have been shown to alleviate this problem. Nevertheless, the superiority of spectral discriminators over spatial discriminators remains unclear. In essence, the spectral discriminator measures the difference between the spectral distribution of real images and generated images. Thus, to understand the effectiveness of the spectral discriminator, a spectral viewpoint is indispensable.

In this paper, we investigate the spectral discriminators in the framework of GAN-based SR, as the quality of the super-resolved images is susceptible to spectral change\footnote[1]{It is well known that downsampling leads to loss or aliasing of frequency components, so SR image quality is susceptible to spectral change.}. Before examining the spectral discriminator, we evaluate how the spatial discriminator behaves in SR. \cref{fig:fig1:a} shows the spectra statistics of Real-ESRNet~\cite{wang2021real}~(distortion-oriented) and Real-ESRGAN~\cite{wang2021real} on the DIV2K~\cite{agustsson2017ntire} dataset, which illustrates that the introduction of the discriminator encourages the generator to produce more high-frequency components. In other words, the generator improves the perceptual quality by matching the spectra of real images\footnote[2]{As the Fourier transform is linear and invertible, the difference between the distributions in the spatial domain and frequency domain is equivalent. Therefore, the closer the distributions are in the frequency domain, the closer they are in the spatial domain.}. Moreover, we estimate the average score of Real-ESRGAN's spatial discriminator on real, low-quality, and generated images~(see \cref{fig:fig1:b}-\cref{fig:fig1:d}); despite the presence of noise and artifacts, low-quality images obtained the highest scores, indicating that the spatial discriminator may favor high-frequency data, even if it is incorrect. A natural question is whether the spectral discriminator's effectiveness is related to its ability to compensate for this high-frequency flaw.

In an effort to answer the question, we further analyzed the spectra and observed that the spectra could be divided into three ranges, as revealed in \cref{fig:fig1:a}. We establish that it is a common phenomenon in image super-resolution algorithms from the frequency perspective of the PD tradeoff, which encourages us to analyze the spatial/spectral discriminator by examining its robustness~\cite{rahaman2019spectral,wang2020high,park2022how,sharma2019effectiveness} under frequency perturbations~\cite{wang2020high,park2022how,sharma2019effectiveness} within these three ranges. We observe that the spectral discriminator indeed performs better than the spatial discriminator in identifying the differences in the \emph{high-frequency} range; however, the spatial discriminator has an advantage in the \emph{low-frequency} range. Therefore, the spatial and spectral discriminators are complementary and are better to be used in combination.

Moreover, taking into account that the previous spectral discriminator~\cite{chen2021ssd,jung2021spectral,schwarz2021frequency,fuoli2021fourier,kim2021unsupervised} is an MLP with 1D reduced spectrum as input, resulting in a loss of input information, we propose spectral Transformer, which calculate patch-wise Fourier transform and then aggregate spectra with Transformer~\cite{dosovitskiy2021an}. Combined with spatial Transformer, our Dual Transformer~(\textbf{DualFormer}) encourage the results of generator achieve better spectra alignment to real images, leading to an improved PD tradeoff. Additionally, we conducted NR-IQA tasks to demonstrate the alignment of our method with human perception. Our method achieved better performance on natural distortion datasets such as LIVE-itW~\cite{ghadiyaram2015massive} and KonIQ-10K~\cite{hosu2020koniq}, as well as on image-processing distortion datasets such as PIPAL~\cite{jinjin2020pipal}, compared to the previous baseline~\cite{zhu2021recycling}.


\section{Related Work}
\label{sec:related_work}
\textbf{The frequency perspective of GAN.} Many studies have examined GAN from a frequency perspective and have reached a consensus that there exist spectral discrepancies between the generator outputs and real images. While most studies attribute these discrepancies to the architecture of the generator~\cite{chandrasegaran2021closer,durall2020watch,frank2020leveraging,gal2021swagan,jiang2021focal,khayatkhoei2022spatial,schwarz2021frequency}, other works point to the discriminator~\cite{chen2021ssd, jung2021spectral,gal2021swagan,schwarz2021frequency}. Concerning the generator, the frequency bias of the upsampling operations was revealed. Specifically, interpolation methods such as bilinear interpolation and nearest interpolation are resistant to generating high frequencies~\cite{chandrasegaran2021closer,durall2020watch,schwarz2021frequency}, whereas zero insertion between pixels introduces excessive high-frequency artifacts~\cite{chandrasegaran2021closer,durall2020watch,schwarz2021frequency,odena2016deconvolution}. Regarding the discriminator, Schwarz~\etal~\cite{schwarz2021frequency} discovered that the spatial discriminator struggles with the low magnitude of the high-frequency content. They experimentally found that the spectral discriminator~\cite{jung2021spectral,chen2021ssd} can reduce the spectral discrepancies, but the misalignment in the high-frequency is not fully addressed. Notably, image fidelity was improved by replacing the input of the spectral discriminator from the reduced spectrum to the full spectrum~\cite{schwarz2021frequency}. However, since the spectral discriminator is an MLP, which does not scale to the real-world setting, we solve this problem by introducing per-patch Fourier Transform, which enables compatibility between Transformer~\cite{dosovitskiy2021an,vaswani2017attention} and frequency domain data.

\textbf{GAN based image super-resolution.} Image super-resolution has seen rapid development~\cite{dong2015image,liang2021swinir,dai2019second,zhang2018residual,gu2021interpreting,lim2017enhanced,wang2020deep,kim2016accurate,zhang2018image,ledig2017photo,pan2022towards} since the introduction of SRCNN~\cite{dong2015image}. Progress has been made on pixel-wise image reconstruction metrics, such as peak signal-to-noise ratio (PSNR). However, these metrics have been shown to have poor correlation with human perception~\cite{ledig2017photo,blau2018perception}. Thus, GAN~\cite{goodfellow2014generative} is leveraged to improve perceptual quality by matching reconstructed and real image distribution~\cite{ledig2017photo}. Following this, a lot of research have improved architecture~\cite{wang2018esrgan,wang2021real,wang2018recovering} or training methods~\cite{sajjadi2017enhancenet,wang2018esrgan,zhang2019ranksrgan,wang2021real,ma2022rectified,park2018srfeat,ma2020structure,fuoli2021fourier}. Like our method, many studies~\cite{fuoli2021fourier,gastineau2021generative} also utilize an additional spectral discriminator. However, they are focused on loss design for efficient SR, and we are driven by the effectiveness problem of the spectral discriminators. Besides answering the above question, we further improve the architecture of the spectral discriminator.

\textbf{Opinion-unaware NR-IQA.} Opinion-unaware NR-IQA~(OU NR-IQA) aims to estimate the perceptual quality of images without access to human-labeled data. There are few studies on deep learning-based OU NR-IQA~\cite{liu2017rankiqa,liu2019exploiting,zhu2021recycling,gu2019no,madhusudana2022image}, and most of them are built on the idea of learning from ranking~\cite{gao2015learning}. To illustrate, Liu~\etal~\cite{liu2017rankiqa,liu2019exploiting} train a siamese Network~\cite{chopra2005learning} to rank images using synthetic distortions for which relative image quality is known. The most relevant work to us is RecycleD~\cite{zhu2021recycling}, where the authors propose that a discriminator trained with adversarial loss can predict perceptual quality, and they show that the discriminator of ESRGAN~\cite{wang2018esrgan} exhibits good IQA performance.


\section{The Frequency Perspective of PD tradeoff}
\label{sec:3}

\cref{fig:fig1:a} shows that the spectra can be roughly divided into three ranges. We argue that it is a common phenomenon of the image SR algorithm through the lens of the frequency perspective of the PD tradeoff.

PD tradeoff~\cite{blau2018perception} claims an image restoration~(\eg, SR) algorithm can be potentially improved only in terms of its perceptual quality or its distortion, one at the expense of the other. Specifically, the loss of an image restoration algorithm can be formulated as

\begin{equation}
	\ell_{\text {gen }}=\lambda_{1}\ell_{\text {distortion }}+\lambda_{2}\ell_{\text {perception}}\text{,}
\end{equation}
where $\ell_{\text {distortion }}$ is usually the $L_1$ loss measuring the per-pixel difference between original and reconstructed images, and $\ell_{\text {perception }}$ is the adversarial loss that measures how close the generator distribution and the real distribution are.

Given the spectral properties of the natural images, the spectra can be divided into three ranges from the frequency perspective of the PD tradeoff:

\textbf{Low-frequency Range.} As the power spectrum of natural images decays exponentially~\cite{schwarz2021frequency}, the natural image mainly comprise low-frequency components. Therefore, the generator will have a high priority to recover low-frequency components since any deviation in this range would cause both high distortion and low perceptual quality. From \cref{fig:fig1:a}, both Real-ESRGAN and Real-ESRNet accurately match real spectra in this spectrum range.

\textbf{Middle-frequency Range.} The distortion metrics are dominated by the difference in the low-frequency range. As demonstrated in \cref{fig:fig1:a}, a generator without the adversarial loss, \eg, Real-ESRNet, will sacrifice the ability to generate higher frequencies to focus on the low-frequency component, resulting in blurred output images. Conversely, equipped with the adversarial loss, Real-ESRGAN is motivated to match real spectra in this range. Additionally, \cref{tab:tab1} shows Real-ESRGAN reaches higher distortion in the low-frequency range and lower distortion in higher frequency ranges compared to Real-ESRNet. In other words, higher perceptual quality is related to lower distortion in middle and high-frequency ranges.

\textbf{High-frequency Range.} The discriminator encourages the generator to generate more components in middle and high-frequency ranges. However, there are subtle yet critical differences between these two ranges. Specifically, the power spectrum of natural images drops suddenly at the highest frequency, resulting in a negligible impact of the distortion term in this range. This is clearly demonstrated by the catastrophic component loss of Real-ESRNet in the high-frequency range of \cref{fig:fig1:a}. Consequently, this range is primarily governed by the perception term. As shown in \cref{fig:fig1:a}, Real-ESRGAN generates random high-frequency details in this range guided by the adversarial loss.

\begin{table}[t]
	\setlength{\abovecaptionskip}{3pt}
	\centering
	\resizebox{0.705\linewidth}{7.05mm}{
		\begin{tabular}{lccc}
			\toprule
			& \multicolumn{1}{c}{Low} & \multicolumn{1}{c}{Middle} & \multicolumn{1}{c}{High} \\
			\midrule
			Real-ESRNet~\cite{wang2021real} & 45.35 & 45.95 & 6.06 \\
			Real-ESRGAN~\cite{wang2021real} & 56.10 & 37.82 & 4.79 \\
			\bottomrule
		\end{tabular}
	}
	\caption{\textbf{Magnitude RMSE for each of the three frequency ranges.} The discriminator mitigates distortion in the middle and high-frequency ranges.}
	\label{tab:tab1}
	\vspace{-1.5em}
\end{table}

\begin{figure*}[t]
	\setlength{\abovecaptionskip}{3pt}
	\centering
	\includegraphics[width=0.8\linewidth]{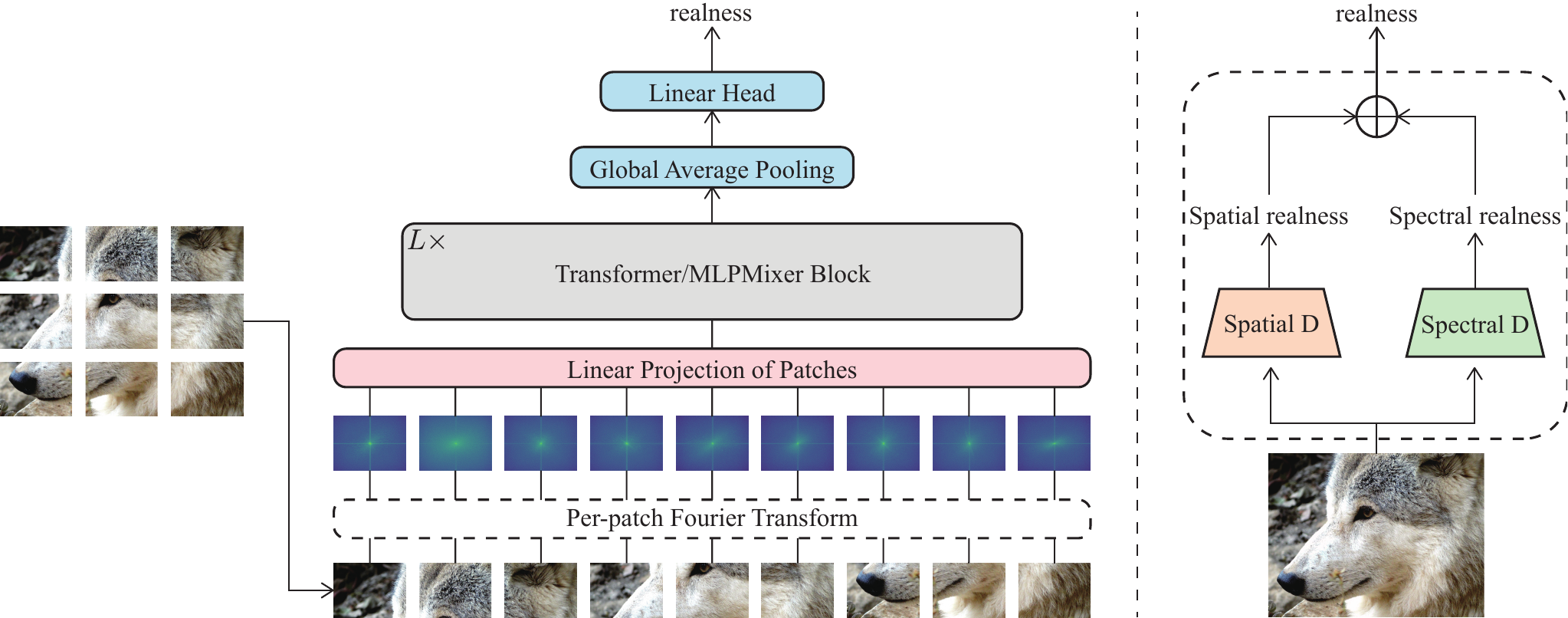}
	\caption{\textbf{Spectral Transformer.} After splitting the image into fixed-size patches, we apply patch-wise Fourier transform on each of them and aggregate the spectra by Transformer. The overall realness is the average of spatial realness and spectral realness.}
	\label{fig:fig2}
	\vspace{-1.5em}
\end{figure*}

\begin{figure}[t]
	\setlength{\abovecaptionskip}{3pt}
	\centering
	\begin{subfigure}[b]{0.495\linewidth}
		\setlength{\abovecaptionskip}{3pt}
		\centering
		\includegraphics[width=\linewidth]{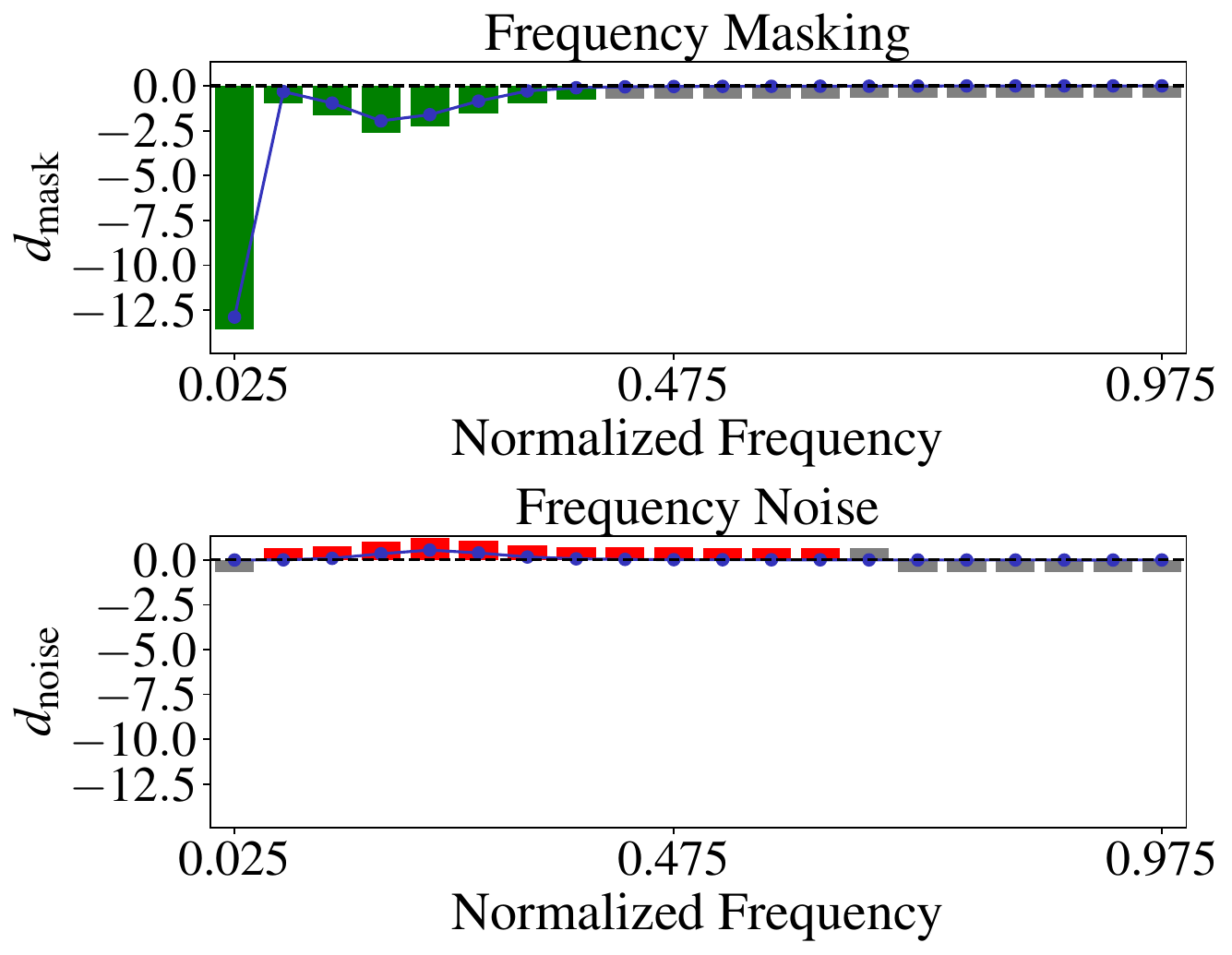}
		\caption{Spatial Transformer}
		\label{fig:fig3:a}
	\end{subfigure}
	\begin{subfigure}[b]{0.495\linewidth}
		\setlength{\abovecaptionskip}{3pt}
		\centering
		\includegraphics[width=\linewidth]{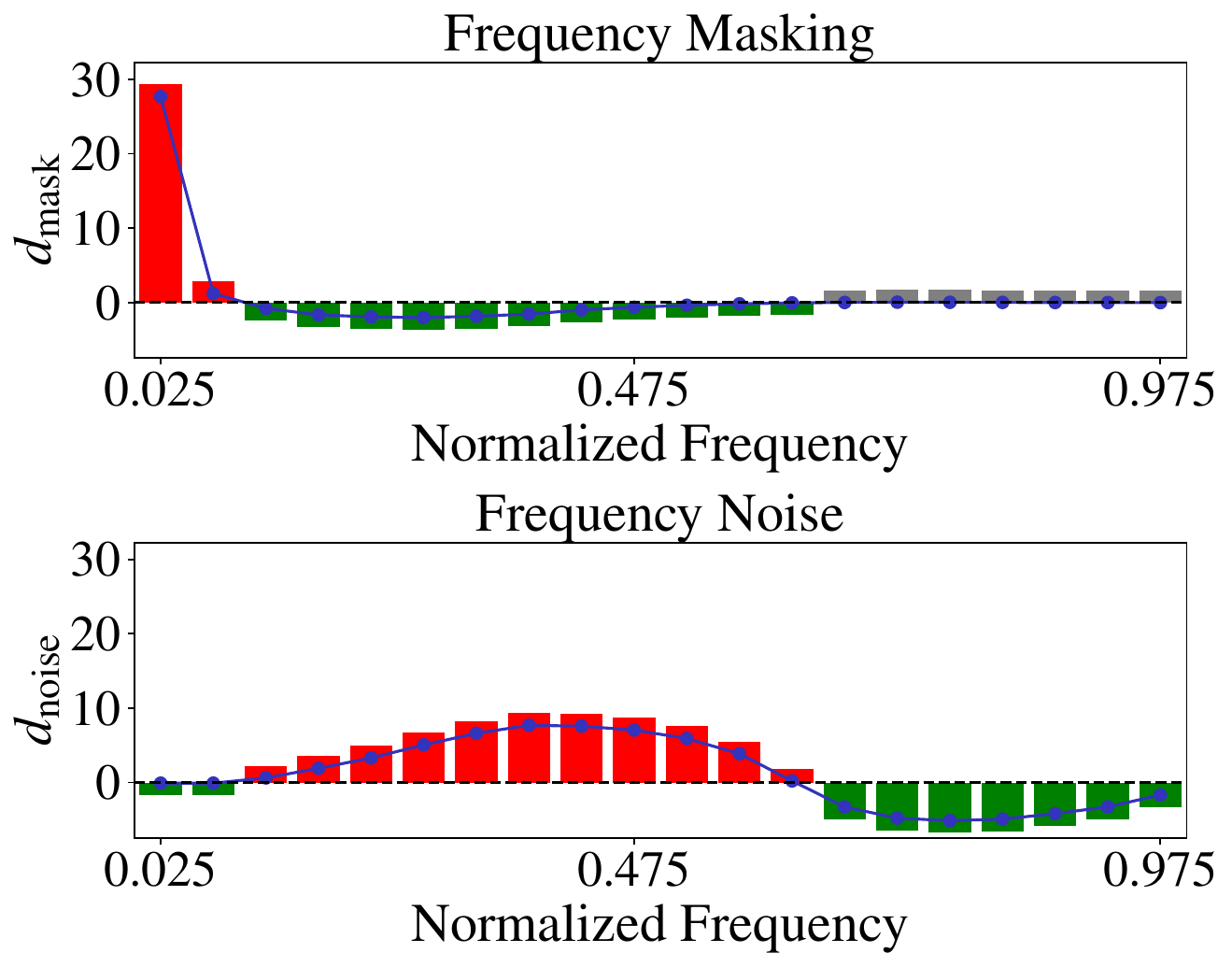}
		\caption{Spectral Transformer}
		\label{fig:fig3:b}
	\end{subfigure}
	\caption{\textbf{Robustness behavior of discriminators under frequency perturbations.} The \textcolor[RGB]{255,0,0}{positive} and \textcolor[RGB]{0,128,0}{negative} values indicate the degree to which the discriminator considers the input image to be real or fake compared to the original image. Negligible differences are shown in \textcolor[RGB]{128,128,128}{gray}. In summary, the spatial discriminator is an expert at discriminating low-frequency masking, while the spectral discriminator is better at discriminating high-frequency noise. Moreover, there is a tradeoff between the discriminator's ability to determine frequency masking and noise in a specific range.}
	\vspace{-1em}
	\label{fig:fig3}
\end{figure}

\section{Method and Analyses}
\label{sec:4}
To compare the spatial and spectral discriminators, a unified architecture that works for both domains is indispensable. But so far, such an architecture does not exist, making it unfair to compare different architectures. To solve this problem, we present our Spectral Transformer in this section, which uses the Transformer in the spectral domain by applying a per-patch Fourier Transform. Then, we analyze the differences between spatial and spectral discriminators from a frequency perspective. Furthermore, we investigate how patch size affects the Spectral Transformer and propose the Dual Transformer as a new discriminator.

\subsection{Spectral Transformer}
\label{sec:4:1}
Several studies have demonstrated that it is possible to improve learning spectral statistics by adding a discriminator on the Fourier spectrum~\cite{schwarz2021frequency,chen2021ssd,jung2021spectral,fuoli2021fourier,kim2021unsupervised}. However, these studies take the reduced spectrum as input because their spectral discriminator adopts the MLP architecture. CNNs do not apply to spectrum data~\cite{schwarz2021frequency}, and if the 2D Fourier spectrum is directly used as input to the MLP, the time and space complexity would be unaffordable. Schwarz~\textit{et al.}~\cite{schwarz2021frequency} showed that replacing the reduced spectrum with the full Fourier transform indeed improves the image fidelity in the low-resolution case. Thus, there is an urgent need for a new architecture that can process full spectrum input efficiently and effectively.

We solve this problem with our Spectral Transformer~(SpecFormer), which harnesses the Transformer by using a per-patch Fourier Transform. First, the image is cropped into patches in the spatial domain according to spatial continuity. Then, we apply Fourier Transform to each patch separately. Finally, these transformed patches are arranged in the original spatial order as input to patch-based networks, \ie, Transformer~\cite{dosovitskiy2021an}~(we particularly call it Spatial Transformer, dubbed SpatFormer, to distinguish from our Spectral Transformer); see \cref{fig:fig2} for an intuitive explanation. Our approach has two benefits. Firstly, SpecFormer takes 2D full spectra, which are powerful and scalable. Secondly, the consistent architecture facilitates our analysis of the differences between the spatial and spectral discriminator in \cref{sec:4:2}.\vspace{-0.05em}

\begin{figure*}[t]
	\setlength{\abovecaptionskip}{3pt}
	\centering
	\begin{subfigure}[b]{0.246\linewidth}
		\setlength{\abovecaptionskip}{3pt}
		\centering
		\includegraphics[width=\linewidth]{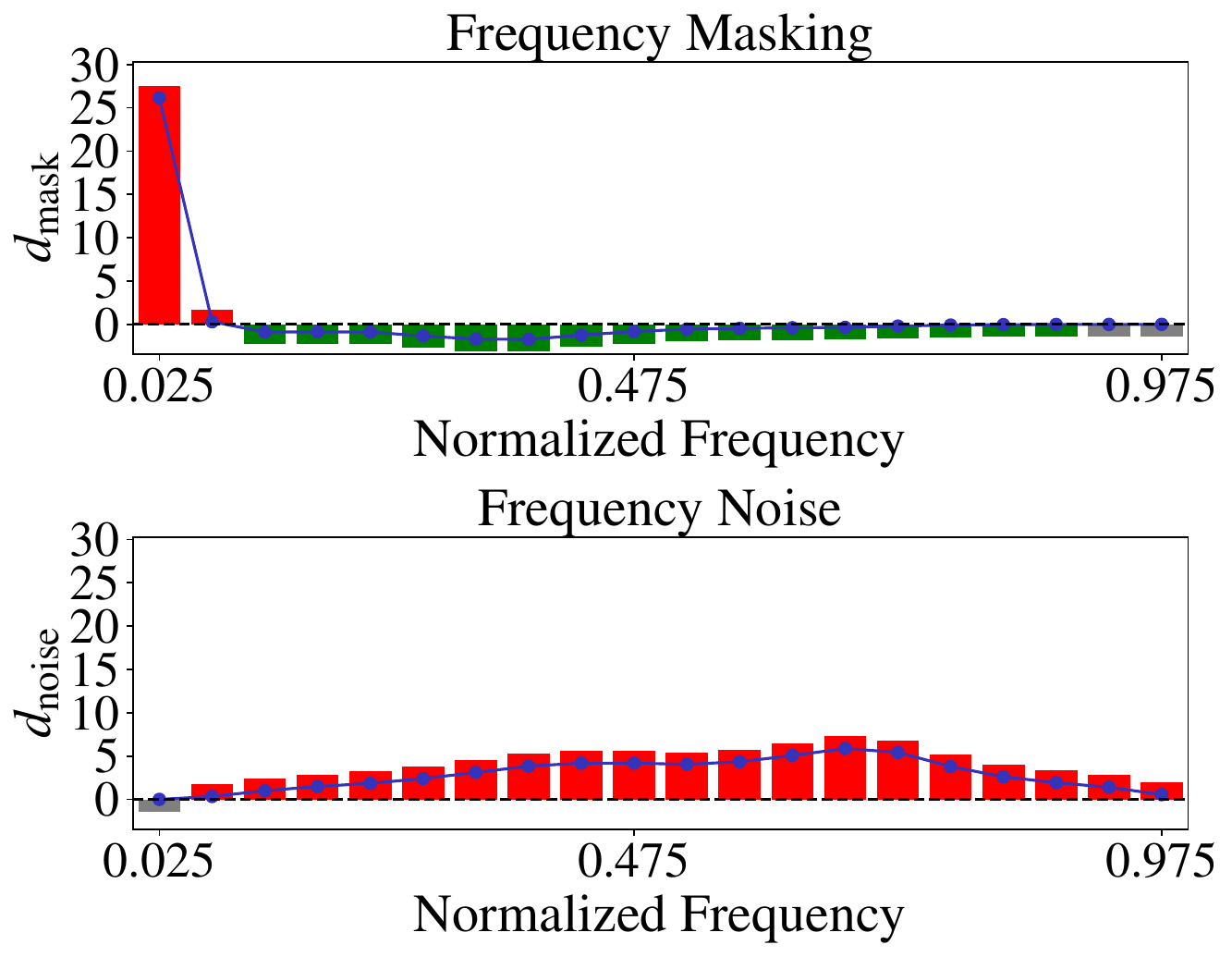}
		\caption{Patch size $8\times 8$}
		\label{fig:fig4:a}
	\end{subfigure}
	\begin{subfigure}[b]{0.246\linewidth}
		\setlength{\abovecaptionskip}{3pt}
		\centering
		\includegraphics[width=\linewidth]{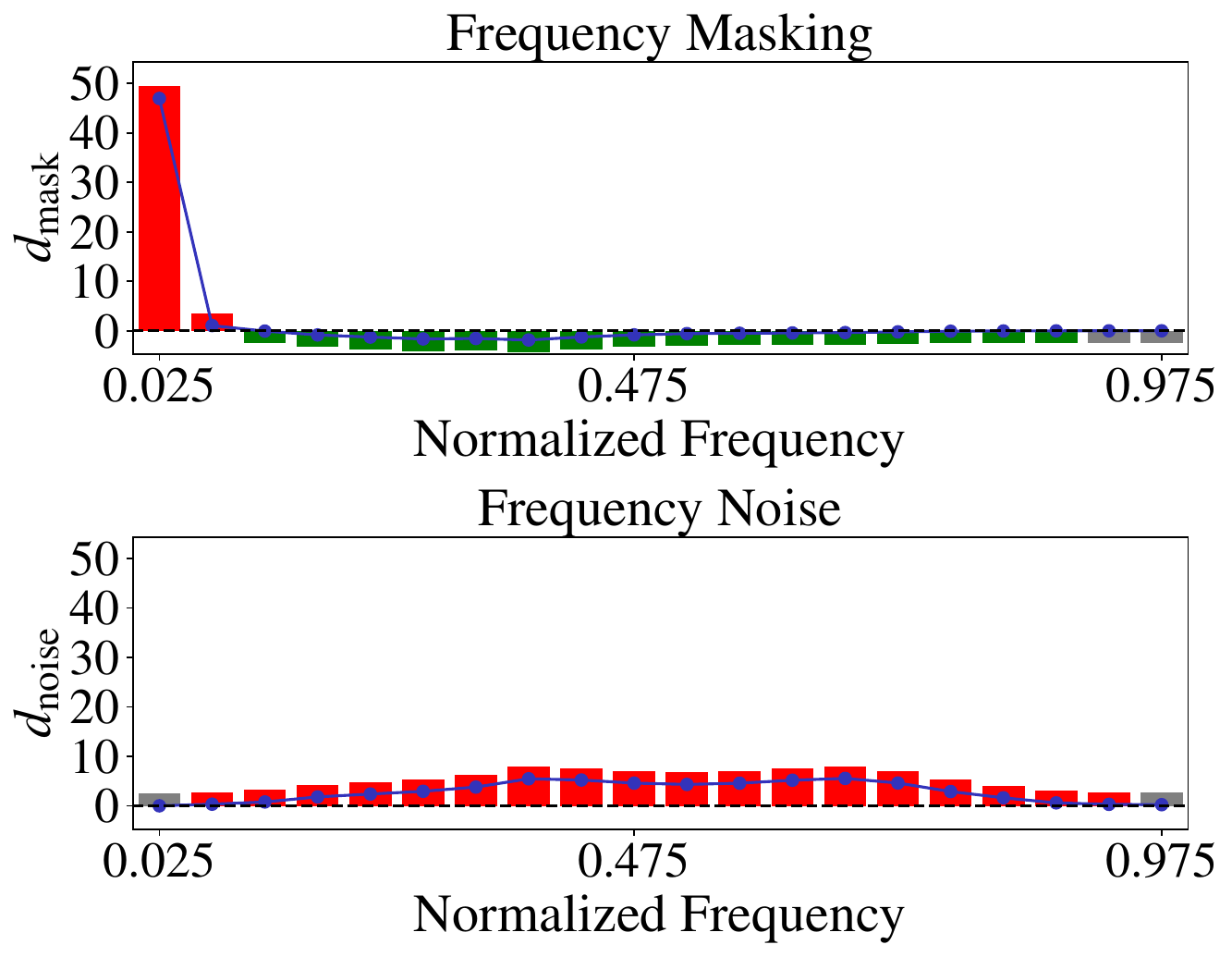}
		\caption{Patch size $16\times 16$}
		\label{fig:fig4:b}
	\end{subfigure}
	\begin{subfigure}[b]{0.246\linewidth}
		\setlength{\abovecaptionskip}{3pt}
		\centering
		\includegraphics[width=\linewidth]{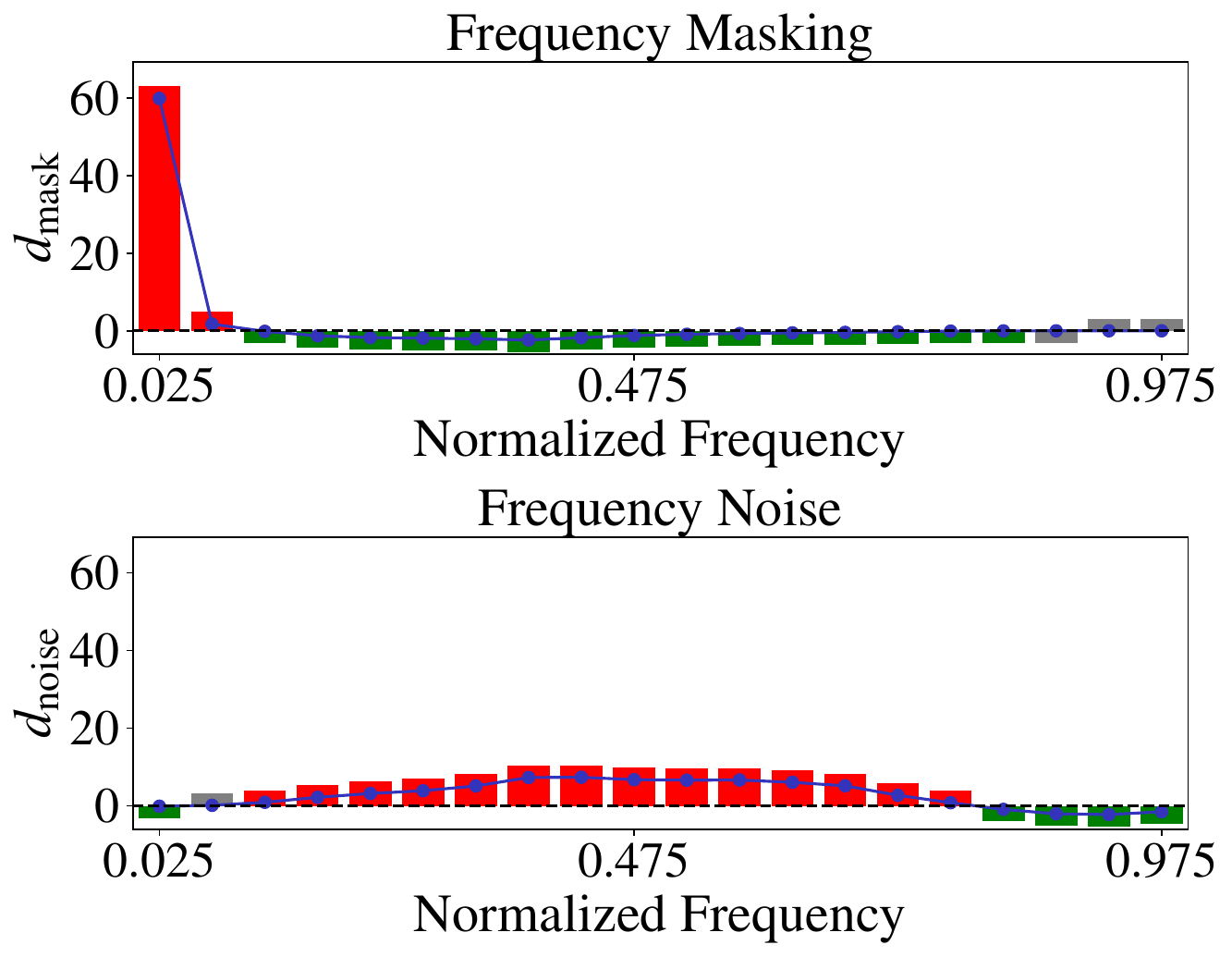}
		\caption{Patch size $32\times 32$}
		\label{fig:fig4:c}
	\end{subfigure}
	\begin{subfigure}[b]{0.246\linewidth}
		\setlength{\abovecaptionskip}{3pt}
		\centering
		\includegraphics[width=\linewidth]{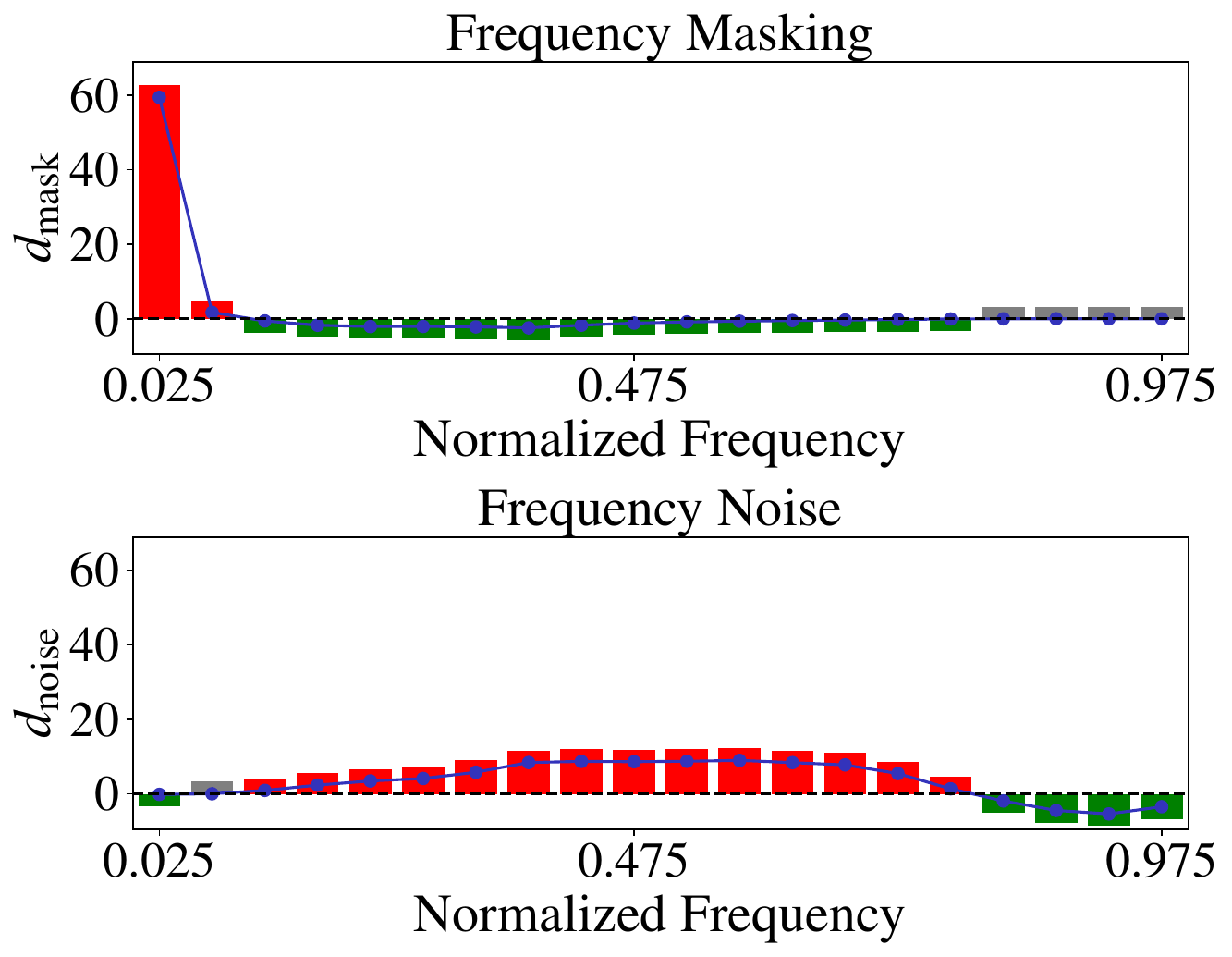}
		\caption{Patch size $64\times 64$}
		\label{fig:fig4:d}
	\end{subfigure}
	\caption{\textbf{Influence of patch size on the robustness behavior of spectral discriminators.} The ability that can discriminate high-frequency noises emerges at a relatively large patch size~\ie $32\times 32$. Consequently, for the Spectral Transformer, a larger patch size can be selected compared to the Spatial Transformer, achieving both effectiveness and computational efficiency.}
	\vspace{-1.3em}
	\label{fig:fig4}
\end{figure*}

\subsection{On the Robustness of Discriminator}
\label{sec:4:2}

The natural division of the spectra encourages us to analyze the discriminator by examining its robustness~\cite{rahaman2019spectral,wang2020high,park2022how,sharma2019effectiveness} under frequency perturbations~\cite{wang2020high,park2022how} in the three spectra ranges. Specifically, we evaluate the robustness of the discriminator in the three ranges under frequency masking~\cite{wang2020high} and noise~\cite{park2022how}. Here we focus on the SpatFormer and SpecFormer, they are same Transformer architecture with or without per-patch Fourier Transform. In addition, we have also studied the behavior of other discriminators and examined how the scaling factory of SR affects the three-range behavior of the discriminator, which further validates the rationality of our analysis method. Please refer to our supplementary for dedicated details.

Denote $\mathbf{I}_{\text {mask}}$ and $\mathbf{I}_{\text {noise}}$ as the image altered by frequency masking and noise. They are defined as follows:
\begin{equation}
	\begin{aligned}
		&\mathbf{I}_{\text {mask }}=\mathcal{F}^{-1}\left(\mathcal{F}\left(\mathbf{I}\right) \odot \overline{\mathbf{M}_f\left(l,r\right)} \right),\\
		&\mathbf{I}_{\text {noise }}=\mathbf{I}+\mathcal{F}^{-1}\left(\mathcal{F}\left(\delta\right) \odot \mathbf{M}_{f}\left(l,r\right)\right),
	\end{aligned}
\end{equation}
where $\mathbf{I}$ is the clean image, $\delta$ is random noise. $\mathcal{F}\left(\cdot\right)$ and $\mathcal{F}^{-1}\left(\cdot\right)$ are Fourier transform and inverse Fourier transform, respectively. $\mathbf{M}_f\left(l,r\right)$ represents the frequency mask, which is $1$ in the frequency interval $\left[l,r\right]$ and $0$ elsewhere. $\overline{\mathbf{M}_f}$ is an element-wise logical inversion of $\mathbf{M}_f$. Namely, for components within the radius interval $\left[l,r\right]$ on the spectrum, we remove them or add noise to them. Then, we monitor the relative difference in the score between altered and real images, $d_\text{mask}=\mathbb{E}\left[D\left(\mathbf{I}_\text{mask}\right)\right]-\mathbb{E}\left[D\left(\mathbf{I}\right)\right]$ and $d_\text{noise}=\mathbb{E}\left[D\left(\mathbf{I}_\text{noise}\right)\right]-\mathbb{E}\left[D\left(\mathbf{I}\right)\right]$. These values indicate how true the discriminator considers it to be compared to the real image. For the purposes of the following analyses, it should be noted that frequency noise is a kind of small difference in a frequency range, while frequency mask corresponds to large deficiency in that range.

We expect the discriminator to always output negative values. However, as shown in \cref{fig:fig3}, both the spatial and spectral discriminators fail in some ranges. In particular, the spectral discriminator exhibits a behavior mirror to that of the generator, characterized by three distinct ranges. This phenomenon can be understood from the frequency perspective of PD tradeoff. 
In the \textbf{low-frequency range}, the generated images closely resemble the ground truth images with only subtle differences. Consequently, the discriminator learns to discriminate subtle differences~(the case of frequency noise), while it cannot discriminate frequency masking. Regarding the \textbf{middle-frequency range}, the generated images exhibit a deficiency state~(lower magnitude than real images in frequency). Since compensating for this deficiency optimizes both the distortion and perceptual quality, the discriminator learns to discriminate this deficiency~(corresponds to the frequency masking) in this range while not recognizing the frequency noise. In the \textbf{high-frequency range}, the distortion term barely affects the image, and the generator tricks the discriminator by generating high-frequency ``noise." Thus, the discriminator only learns to discriminate frequency noise. In conclusion, the discriminator learns based on what the generator produces, and due to the fact generator has different generation inclinations in each range, the discriminator also learns different abilities in each range. It is noteworthy that the spatial discriminator does not directly operate in the spectral domain, thereby we cannot analyze it using the same approach.

Finally, as shown in \cref{fig:fig3:a}-\cref{fig:fig3:b}, the application of the same architecture to spatial and frequency domains separately results in significant differences, which illustrates that \emph{the spatial discriminator is an expert at discriminating deficiency in the low-frequency range, while the spectral discriminator is better at discriminating noise in the high-frequency range}. Thus, the spatial and spectral discriminators are complementary and should be used in combination.\vspace{-0.1em}

\subsection{How patch size affects Transformers}
\label{sec:4:3}
The choice of patch size has a crucial impact on the efficiency and effectiveness of Transformers. While small patch sizes may improve performance of SpatFormers, it results in lower efficiency~\cite{dosovitskiy2021an}. However, this may not hold true for SpecFormers, since applying Fourier transform on small patches may not be meaningful. To verify this, we investigate the effect of patch size on the robustness behavior of SpecFormer. As shown in \cref{fig:fig4}, the SpecFormer requires a patch size of at least $32\times 32$ to effectively distinguish high-frequency noise. While we conducted a similar analysis for SpatFormer, its robustness behavior is not sensitive to patch size. In light of above observations, we can select a larger patch size for the SpecFormer, which yields better performance and is also more efficient.

\subsection{Dual Transformer}
Based on our previous analysis, we have found that the Spatial discriminator and Spectral discriminator possess complementary discriminative power. Therefore, we propose a novel discriminator called Dual Transformer~(DualFormer) by combining the Spatial Transformer and Spectral Transformer. In particular, considering that the Spectral Transformer exhibits a preference for larger patch sizes, we use a Spectral Transformer with a minimum patch size of $32\times 32$ in our work. For implementation, we utilize ViT~\cite{dosovitskiy2021an} as the basis, adopt relative position encoding, and apply global average pooling for realness prediction. Our configuration is lightweight, with only 10 blocks and a hidden layer dimension of 96. We have tested heavier networks, but they did not exhibit any improvement in performance.

As we use two Transformers as discriminators, the efficiency of our method may become a concern. In fact, our discriminator is more efficient than commonly used VGG~\cite{wang2018esrgan} and U-Net~\cite{wang2021real}~(in terms of parameters, FLOPs and activations, see supplementary for details), since we use large patch sizes and small layer dimension.\vspace{-0.1em}

\section{Experimental Results}
We validate the proposed method from two perspectives. Firstly, we verify whether our method can help the generator to better match spectra and achieve better SR performance. Secondly, we consider whether our method improves the discriminator's ability to predict image quality from the perspective of OU NR-IQA~\cite{zhu2021recycling}.

\begin{table*}[t]
	\centering
	\resizebox{\linewidth}{22mm}{
		\begin{tabular}{ccccccccccc}
			\toprule
			Dataset & Metric & SRGAN~\cite{ledig2017photo} &  RankSRGAN~\cite{zhang2019ranksrgan} & ESRGAN~\cite{wang2018esrgan} & SPSR~\cite{ma2020structure} & ESRGAN + LDL~\cite{liang2022details}& FFTGAN~\cite{fuoli2021fourier} & Ours  \\
			\midrule
			\multirow{3}{*}{BSD100} & $\uparrow$PSNR & 25.5544 & 25.5315 & 25.3726 & 25.7170 & 25.5685 & \textcolor[RGB]{0,0,255}{25.7900} & \textcolor[RGB]{255,0,0}{26.4826} \\
			& $\uparrow$SSIM & 0.6542 & 0.6518 & 0.6515 & \textcolor[RGB]{0,0,255}{0.6658} & 0.6616 & 0.6580 & \textcolor[RGB]{255,0,0}{0.6895} \\
			& $\downarrow$LPIPS & 0.1783 & 0.1772  & 0.1597 & \textcolor[RGB]{255,0,0}{0.1558} & 0.1592 & 0.1580 & \textcolor[RGB]{0,0,255}{0.1573} \\
			\midrule
			\multirow{3}{*}{DIV2K} & $\uparrow$PSNR & 28.1646 & 28.0916 & 28.0465 & 28.3978 & 28.2440 & \textcolor[RGB]{0,0,255}{28.6300} & \textcolor[RGB]{255,0,0}{29.3049}  \\
			& $\uparrow$SSIM & 0.7745 & 0.0646 & 0.7669 & \textcolor[RGB]{0,0,255}{0.7821} & 0.7758 & 0.7800 & \textcolor[RGB]{255,0,0}{0.8023} \\
			& $\downarrow$LPIPS & 0.1257 & 0.1239 & 0.1142 & \textcolor[RGB]{0,0,255}{0.1069} & 0.1133 & 0.1130 & \textcolor[RGB]{255,0,0}{0.1030} \\
			\midrule
			\multirow{3}{*}{Urban100} & $\uparrow$PSNR & 24.4056 & 24.4233 & 24.6287 & 24.8393 & 24.4777 & \textcolor[RGB]{0,0,255}{25.0500} & \textcolor[RGB]{255,0,0}{25.6870} \\
			& $\uparrow$SSIM & 0.7298 & 0.7265 & 0.7411 & \textcolor[RGB]{0,0,255}{0.7493} & 0.7389 & 0.7380 & \textcolor[RGB]{255,0,0}{0.7727} \\
			& $\downarrow$LPIPS & 0.1439 & 0.1438 & 0.1240 & \textcolor[RGB]{0,0,255}{0.1174} & 0.1243 & 0.1200 & \textcolor[RGB]{255,0,0}{0.1147} \\
			\bottomrule
		\end{tabular}
	}
	\caption{Quantitative comparison of GAN-based SR methods on $\times 4$ super-resolution.}
	\label{tab:tab2}
	\vspace{-1em}
\end{table*}

\begin{figure*}[h]
	\setlength{\abovecaptionskip}{0.4pt}
	\centering
	\includegraphics[width=\linewidth]{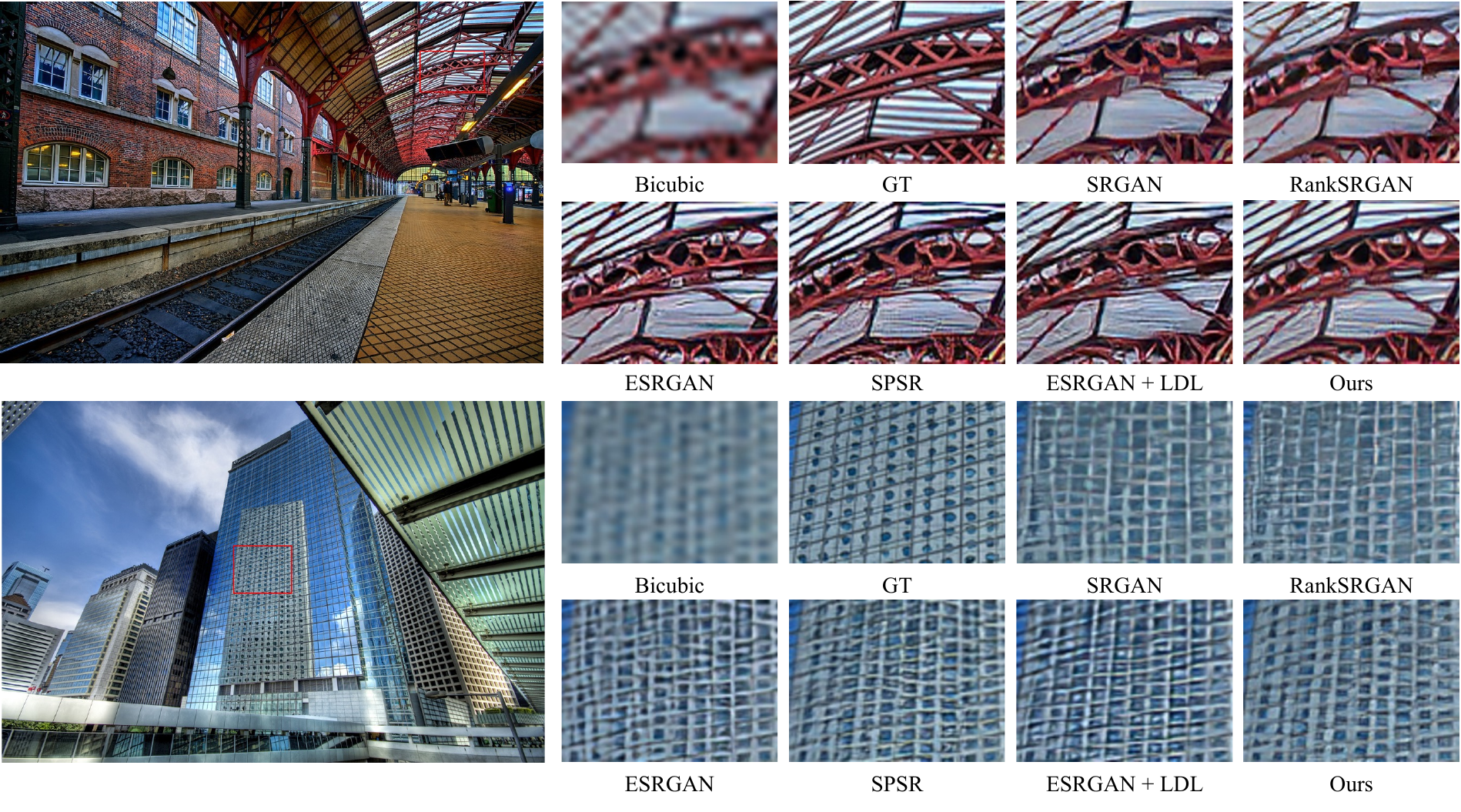}
	\caption{Visual comparison of GAN-based SR methods on $\times 4$ super-resolution.}
	\label{fig:fig5}
	\vspace{-1em}
\end{figure*}

\subsection{Image Super-Resolution}
\textbf{Experimental Setup.} To construct our model, we select ESRGAN~\cite{ledig2017photo} as the baseline and replace its discriminator with ours. The patch size of both SpatFormer and SpecFormer are set to $32\times 32$. The training settings are aligned with ESRGAN. We adopt DF2K~\cite{agustsson2017ntire,lim2017d} as the training set. The training pairs are generated by bicubic downsampling with a scaling factor of 4. The HR patch size is set to 128, and the total batch size is set to 16. Networks are trained with 1000k using only $L_1$ loss and then 400k iterations with a combination of $L_1$ loss, perceptual loss~\cite{wang2021real}, and GAN loss. The loss weights are kept same as ESRGAN.

\textbf{Test sets and metrics.} We evaluate our methods on three datasets: BSD100~\cite{martin2001database}, DIV2K validation set~\cite{agustsson2017ntire}, and Urban100~\cite{huang2015single}. We used PSNR and SSIM~\cite{wang2004image} as distortion metrics and LPIPS~\cite{zhang2018unreasonable} for evaluating perceptual quality.

\begin{figure*}[t]
	\setlength{\abovecaptionskip}{1pt}
	\centering
	\includegraphics[width=\linewidth]{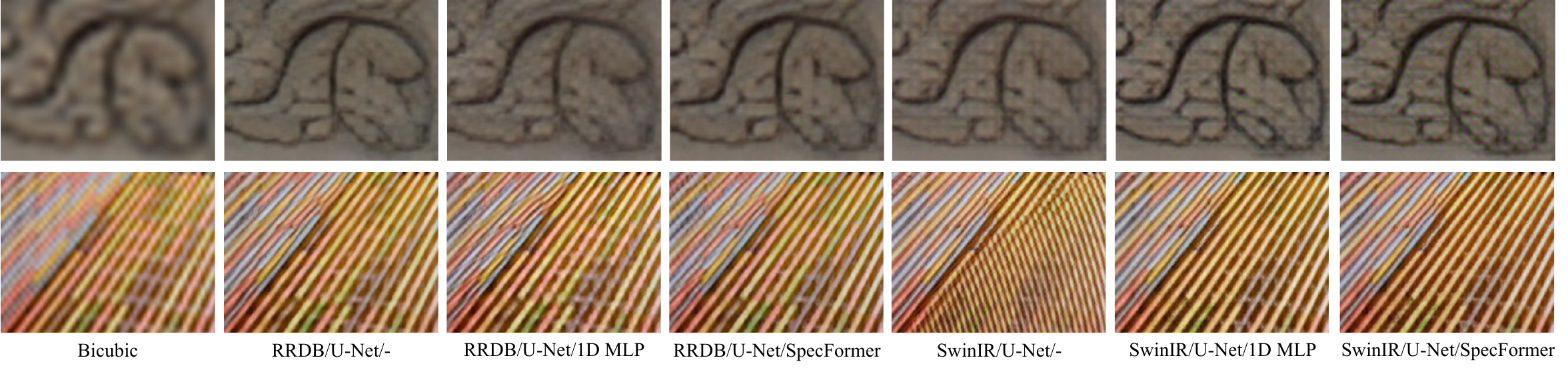}
	\caption{Visual comparisons of various combinations~(generator/spatial discriminator/spectral discriminator) on $\times 4$ super-resolution.}
	\label{fig:fig6}
	\vspace{-1.5em}
\end{figure*}

\textbf{Comparison with state-of-the-arts.} To validate the effectiveness of our method, besides ESRGAN, we compare it with several state-of-the-art~(SOTA) GAN-based SR methods: RankSRGAN~\cite{zhang2019ranksrgan}, SPSR~\cite{ma2020structure}, ESRGAN + LDL~\cite{liang2022details}, FFTGAN~(ESRGAN version)~\cite{fuoli2021fourier}, where RankSRGAN is constructed from SRGAN~\cite{ledig2017photo}, other methods are based on ESRGAN. SPSR improves the architecture of the generator, while other studies, including ours, use the original generator of SRGAN/ESRGAN. For all methods except FFTGAN, we retrain the model using the code provided by the authors to keep the training settings the same as their basis, \ie, SRGAN/ESRGAN. Since the code of FFTGAN is not publicly available, we take the results directly from their paper~(notice that FFTGAN uses a large patch size of $256 \times 256$). \cref{tab:tab2} shows the quantitative comparisons. Overall, our method achieves the best distortion~(PSNR/SSIM) and perceptual quality~(LPIPS) compared to other SOTA methods on all benchmarks. Our discriminator improves the performance of ESRGAN significantly due to its better discriminating ability. ~\cref{fig:fig5} shows the visual comparison, demonstrating that our method produces cleaner results with fewer artifacts than other methods since the spectral discriminator can identify high-frequency noise effectively.

\begin{figure}[t]
	\centering
	\includegraphics[width=0.9\linewidth]{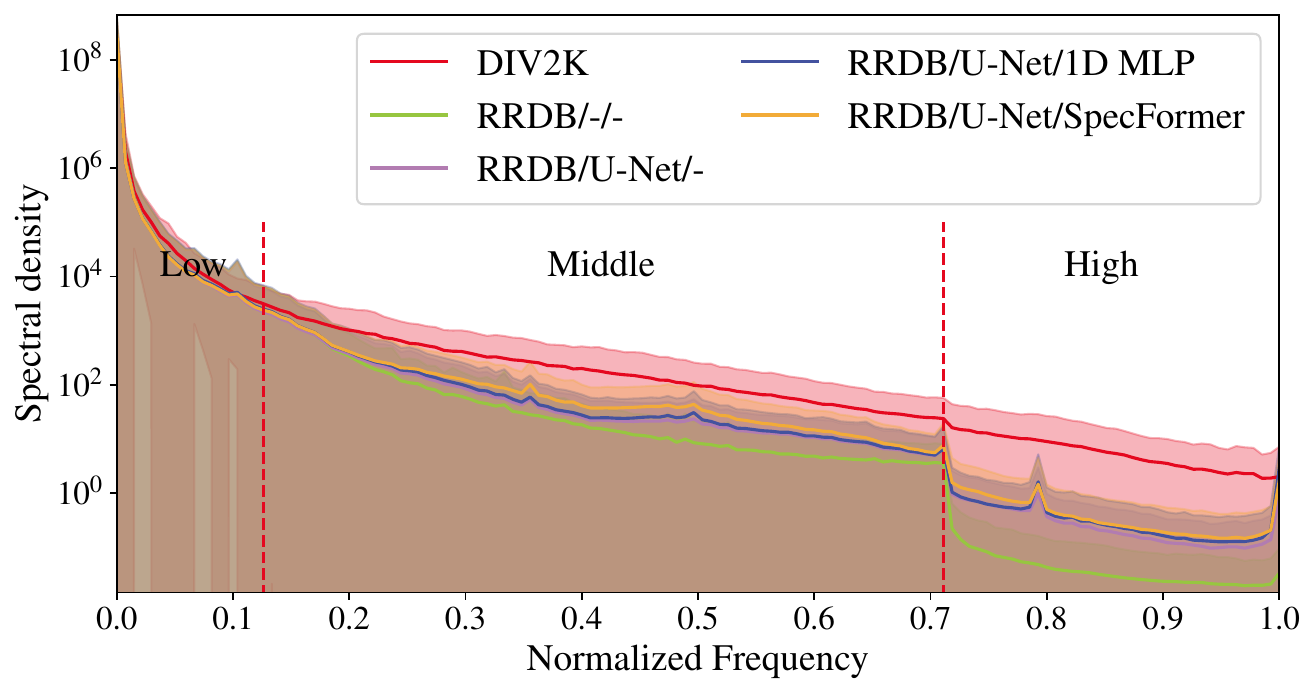}
	\caption{Our method exhibits better spectra alignment.}
	\label{fig:fig7}
	\vspace{-1em}
\end{figure}

\begin{table}[t]
	\setlength{\abovecaptionskip}{3pt}
	\centering
	\resizebox{0.85\linewidth}{18mm}{
		\begin{tabular}{ccccc}
			\toprule
			Method & $\uparrow$PSNR & $\uparrow$SSIM & $\downarrow$LPIPS \\
			\midrule
			ESRGAN & 25.7327 & 0.7144 & 0.2161 \\
			ESRGAN + LDL & 25.6998 & 0.7097 & 0.2215 \\
			SPSR & 25.8033 & 0.7137 & 0.2160 \\
			RankSRGAN & 25.7888 & 0.6953 & 0.2488 \\
			BebyGAN~\cite{li2022best} & 26.3378 & 0.7270 & 0.2077 \\
			\midrule
			Ours~(ESRGAN version) & \textcolor[RGB]{0,0,255}{26.8768} & \textcolor[RGB]{0,0,255}{0.7451} & \textcolor[RGB]{0,0,255}{0.2039}  \\
			Ours~(BebyGAN version) & \textcolor[RGB]{255,0,0}{27.4823} & \textcolor[RGB]{255,0,0}{0.7583} & \textcolor[RGB]{255,0,0}{0.1993} \\
			\bottomrule
		\end{tabular}
	}
	\caption{Comparison under hard gated degradation~(DIV2K).}
	\label{tab:tab3}
	\vspace{-1em}
\end{table}

\textbf{Complex Degradation.} In the preceding, we demonstrated the effectiveness of the spectral discriminator on bicubic degradation. However, as bicubic degradation is quite simple, ESRGAN can already match the real spectra well~(see supplementary for details). To see the further potential of our method, we examine whether our discriminator can improve SRGAN models under a more complex degradation model~(the hard gated degradation model~\cite{zhang2022closer}). In addition to ESRGAN, we also investigate whether our DualFormer can improve BebyGAN~\cite{li2022best} in this case. From \cref{tab:tab3} it can be seen that our method once again shows superior performance, while achieving the best trade-off between distortion and perceptual quality.

\textbf{Ablation study.} We conduct ablation studies to investigate the role of the Spectral Transformer by selecting two representative networks, \ie, RRDB~\cite{wang2018esrgan, wang2021real} and SwinIR~\cite{liang2021swinir}, and evaluating the influence of various combinations of spatial and spectral discriminators. The experiments used a batch size of 32 and a patch size of 192, and were trained for 50,000 iterations under hard gated degradation. \cref{tab:tab4} and \cref{fig:fig6} showed that the Spectral Transformer helped improve the visual quality of the generators. Interestingly, when the Spatial Transformer served as the Spatial Discriminator, the Spectral Transformer not only improved visual quality but also helped reduce distortion. This phenomenon did not occur with other spatial discriminators. These findings suggest that the Spatial Transformer and Spectral Transformer complement each other well. Additionally, we plotted the spectrum profile of the RRDB with different discriminator combinations in \cref{fig:fig7}, demonstrating that spectral discriminators promote high-frequency components and our SpecFormer aligns spectra better.

\begin{table}[t]
	\centering
	\resizebox{\linewidth}{31mm}{
		\begin{tabular}{ccccccc}
			\toprule
			G & Spatial D & Spectral D & $\uparrow$PSNR & $\uparrow$SSIM & $\downarrow$LPIPS \\
			\midrule
			RRDB & VGG & - & 27.0408 & 0.7618 & 0.2359  \\
			RRDB & VGG & 1D MLP~\cite{jung2021spectral} & 26.4706 & 0.7448 & \textcolor[RGB]{0,0,255}{0.2317} \\
			RRDB & VGG & SpecFormer & 26.8322 & 0.7575 & \textcolor[RGB]{255,0,0}{0.2271} \\
			\midrule
			RRDB & SpatFormer & - & 26.9438 & 0.7553 & \textcolor[RGB]{0,0,255}{0.2326} \\
			RRDB & SpatFormer & 1D MLP & 27.0769 & 0.7619 & 0.2493  \\
			RRDB & SpatFormer & SpecFormer & \textbf{27.1460} & \textbf{0.7590} & \textcolor[RGB]{255,0,0}{0.2284} \\
			\midrule
			RRDB & U-Net & - & 26.9435 & 0.7585 & \textcolor[RGB]{0,0,255}{0.2341} \\
			RRDB & U-Net & 1D MLP & 25.9467 & 0.7582 & 0.2358 \\
			RRDB & U-Net & SpecFormer & 26.2829 & 0.7547 & \textcolor[RGB]{255,0,0}{0.2258}  \\
			\midrule
			RRDB & - & 1D MLP & 26.4098 & 0.7549 & \textcolor[RGB]{0,0,255}{0.2331}   \\
			RRDB & - & SpecFormer & 26.4157 & 0.7544 & 0.2319  \\
			\midrule
			SwinIR & U-Net & - & 27.2959 & 0.7593 & \textcolor[RGB]{0,0,255}{0.2387} \\
			SwinIR & U-Net & 1D MLP & 26.5903 & 0.7611 & 0.2447 \\
			SwinIR & U-Net & SpecFormer & 26.6090 & 0.7529 & \textcolor[RGB]{255,0,0}{0.2216}  \\
			\bottomrule
		\end{tabular}
	}
	\caption{Quantitative comparison of various combinations on $\times 4$ super-resolution. Metrics are evaluated on DIV2K validation dataset.}
	\label{tab:tab4}
	\vspace{-1em}
\end{table}

\subsection{No-Reference Image Quality Assessment}

\textbf{Experimental Setup.} Following RecycleD~\cite{zhu2021recycling}, we choose DF2K~\cite{agustsson2017ntire,lim2017d} and OutdoorSceneTraining~\cite{wang2018recovering} as the training dataset. The generator is first trained with the $L_1$ loss for 500 iterations, and then the discriminator is introduced to train another 50k iterations. The HR patch size is set to 192, and the total batch size is set to 16. We report the result of the best-performing checkpoint for each method. For all experiments, the low-resolution images are generated by the simple gated degradation Model~\cite{zhang2022closer}. The patch sizes of SpatFormer and SpecFormer are set to $8\times 8$ and $64\times 64$ respectively.

\textbf{Test sets and metrics.} We evaluate our spectral discriminator on three representative IQA datasets, \ie LIVE-itW~\cite{ghadiyaram2015massive}, KonIQ-10k~\cite{hosu2020koniq} and PIPAL~\cite{jinjin2020pipal}. LIVE-itW and KonIQ-10k contain diverse, authentic distortions, while PIPAL constitutes with many image processing artifacts including the results of perceptual-oriented algorithms. We report three widely used metrics: Pearson linear correlation coeﬃcient~(PLCC), Spearman rank order correlation coeﬃcient~(SRCC), and Kendall rank order correlation coeﬃcient~\cite{kendall1938new}~(KRCC).

\begin{table}[t]
	\centering
	\resizebox{\linewidth}{20mm}{
		\begin{tabular}{ccccccc}
			\toprule
			\multirow{2}{*}{Method} & \multicolumn{3}{c}{LIVE-itW} & \multicolumn{3}{c}{KonIQ-10k} \\
			& $\uparrow$PLCC & $\uparrow$SRCC & $\uparrow$KRCC & $\uparrow$PLCC & $\uparrow$SRCC & $\uparrow$KRCC \\
			\midrule
			QAC~\cite{xue2013learning} & 0.2720 & 0.0457 & 0.0370 & 0.3719 & 0.3397 & 0.2302 \\
			LPSI~\cite{wu2015highly} & 0.2877 & 0.0834 & 0.0524 & 0.2066 & 0.2239 & 0.1504 \\
			IL-NIQE~\cite{zhang2015feature} & 0.5039 & 0.4394 & 0.2985 & 0.5316 & 0.5057 & 0.3504 \\
			SNP-NIQE~\cite{liu2019unsupervised} & \textcolor[RGB]{255,0,0}{0.5201} & \textcolor[RGB]{0,0,255}{0.4657} & \textcolor[RGB]{0,0,255}{0.3162} & \textcolor[RGB]{0,0,255}{0.6340} & \textcolor[RGB]{0,0,255}{0.6285} & \textcolor[RGB]{255,0,0}{0.4435} \\
			\midrule
			dipIQ~\cite{ma2017dipiq} & 0.3180 & 0.1774 & 0.1207 & 0.4429 & 0.2375 & 0.1594 \\
			RankIQA~\cite{liu2017rankiqa} & 0.4528 & 0.4307 & 0.2945 & 0.5028 & 0.4983 & 0.3448 \\
			\midrule
			RecycleD~\cite{zhu2021recycling} & - & - & - & 0.6105 & 0.6020 & 0.4201 \\
			Ours & \textcolor[RGB]{0,0,255}{0.5068} & \textcolor[RGB]{255,0,0}{0.4897} & \textcolor[RGB]{255,0,0}{0.3312} & \textcolor[RGB]{255,0,0}{0.6543} & \textcolor[RGB]{255,0,0}{0.6321} & \textcolor[RGB]{255,0,0}{0.4435} \\
			\bottomrule
		\end{tabular}
	}
	\caption{Quantitative comparison on NR-IQA.}
	\label{tab:tab5}
	\vspace{-1.5em}
\end{table}

\begin{table}[ht]
	\centering
	\resizebox{\linewidth}{25mm}{
		\begin{tabular}{cccccc}
			\toprule
			Spatial D & Spectral D & Traditional & PSNR & GAN & Full \\
			& & SR & based SR & based SR &\\
			\midrule
			VGG~\cite{zhu2021recycling} & - & \textcolor[RGB]{255,0,0}{0.3902} & \textcolor[RGB]{255,0,0}{0.4630} & \textcolor[RGB]{255,0,0}{0.0772} & \textcolor[RGB]{0,0,255}{0.2842} \\
			VGG & 1D MLP~\cite{jung2021spectral} & 0.2692 & 0.3444 & 0.0488 & 0.2146 \\
			VGG & SpecFormer & \textcolor[RGB]{0,0,255}{0.3841} & \textcolor[RGB]{0,0,255}{0.4609} & \textcolor[RGB]{0,0,255}{0.0636} & \textcolor[RGB]{255,0,0}{0.2889} \\
			\midrule
			U-Net  & - & -0.0048 & 0.0354 & 0.0035 & 0.0162 \\
			U-Net & 1D MLP & 0.0144 & 0.0379 & 0.0189 & 0.0477 \\
			U-Net & SpecFormer & \textcolor[RGB]{255,0,0}{0.3700} & \textcolor[RGB]{255,0,0}{0.4391} & \textcolor[RGB]{255,0,0}{0.0598} & \textcolor[RGB]{255,0,0}{0.2936} \\
			\midrule
			- & 1D MLP & 0.0368 & 0.0772 & 0.0301 & 0.0423 \\
			- & SpecFormer & \textcolor[RGB]{255,0,0}{0.3626} & \textcolor[RGB]{255,0,0}{0.4346} & \textcolor[RGB]{255,0,0}{0.0538} & \textcolor[RGB]{255,0,0}{0.2883} \\
			\midrule
			SpatFormer & - & \textcolor[RGB]{255,0,0}{0.3689} & \textcolor[RGB]{0,0,255}{0.4196} & \textcolor[RGB]{0,0,255}{0.0640} & \textcolor[RGB]{0,0,255}{0.2664} \\
			SpatFormer & 1D MLP & 0.2636 & 0.373 & 0.0364 & 0.2301 \\
			SpatFormer & SpecFormer & \textcolor[RGB]{0,0,255}{0.3578} & \textcolor[RGB]{255,0,0}{0.4285} & \textcolor[RGB]{255,0,0}{0.0647} & \textcolor[RGB]{255,0,0}{0.2787} \\
			\bottomrule
		\end{tabular}
	}
	\caption{Comparison~($\uparrow$SRCC) on sub-types of PIPAL training set.}
	\label{tab:tab6}
	\vspace{-1em}
\end{table}

\textbf{Comparison with state-of-the-arts.} We compare our method to representative OU NR-IQA methods: QAC~\cite{xue2013learning}, LPSI~\cite{wu2015highly}, IL-NIQE~\cite{zhang2015feature}, SNP-NIQE~\cite{liu2019unsupervised}, dipIQ~\cite{ma2017dipiq}, RankIQA~\cite{liu2017rankiqa}, and RecycleD~\cite{zhu2021recycling}. Where QAC, LPSI, IL-NIQE, SNP-NIQE are conventional methods, dipIQ, RankIQA, RecycleD are neural network-based methods. Our method is directly based RecycleD, with an extra spectral discriminator. Notably, RecycleD uses sophisticated weighting strategy to get better results, we disable it for simplicity as our target is not performance but to validate the effectiveness of the spectral discriminator to predict perceptual quality.
\cref{tab:tab5} shows the results, our method reached the best results on two datasets, which verified our ensembled discriminator could predict perceptual quality better than the spatial discriminator used in RecycleD.

\begin{table}[t]
	\centering
	\resizebox{0.9\linewidth}{24mm}{
		\begin{tabular}{ccccc}
			\toprule
			Spatial D & Spectral D & $\uparrow$PLCC & $\uparrow$SRCC & $\uparrow$KRCC \\
			\midrule
			VGG~\cite{zhu2021recycling} & - &  \textcolor[RGB]{0,0,255}{0.6240} & \textcolor[RGB]{0,0,255}{0.6034} & \textcolor[RGB]{0,0,255}{0.4200} \\
			VGG & 1D MLP~\cite{jung2021spectral} & 0.5215 & 0.5013 & 0.3434 \\
			VGG & SpecFormer & \textcolor[RGB]{255,0,0}{0.6543} & \textcolor[RGB]{255,0,0}{0.6321} & \textcolor[RGB]{255,0,0}{0.4434} \\
			\midrule
			U-Net  & - & \textcolor[RGB]{0,0,255}{0.4144} & \textcolor[RGB]{0,0,255}{0.3458} & \textcolor[RGB]{0,0,255}{0.2328} \\
			U-Net & 1D MLP & 0.3831 & 0.3021 & 0.2016 \\
			U-Net & SpecFormer & \textcolor[RGB]{255,0,0}{0.6459} & \textcolor[RGB]{255,0,0}{0.6207} & \textcolor[RGB]{255,0,0}{0.4355} \\
			\midrule
			- & 1D MLP & 0.2863 & -0.2339 & -0.1568 \\
			- & SpecFormer & \textcolor[RGB]{255,0,0}{0.6357} & \textcolor[RGB]{255,0,0}{0.6094} & \textcolor[RGB]{255,0,0}{0.4261} \\
			\midrule
			SpatFormer & - & \textcolor[RGB]{0,0,255}{0.6372} & \textcolor[RGB]{0,0,255}{0.6124} & \textcolor[RGB]{0,0,255}{0.4275} \\
			SpatFormer & 1D MLP & 0.575 & 0.5566 & 0.3847 \\
			SpatFormer & SpecFormer & \textcolor[RGB]{255,0,0}{0.6394} & \textcolor[RGB]{255,0,0}{0.6220} & \textcolor[RGB]{255,0,0}{0.4360} \\
			\bottomrule
		\end{tabular}
	}
	\caption{Quantitative comparison of various combinations on NR-IQA. Metrics are evaluated on KonIQ-10k dataset.}
	\label{tab:tab7}
	\vspace{-1.5em}
\end{table}

\textbf{Ablation study.} We conducted experiments to investigate the impact of SpecFormer on various spatial discriminators. As shown in \cref{tab:tab6}, in most cases, using our SpecFormer achieved the best results. Specifically, when the spatial discriminator was VGG, SpecFormer improved the performance on the complete PIPAL dataset, but showed a slight decrease in performance on several SR-related subsets. Additionally, all combinations performed poorly in GAN-based SR, which is consistent with the results of Zhu~\etal~\cite{zhu2021recycling}. Results on the KonIQ-10K dataset are shown in \cref{tab:tab7}, and our method achieved the best performance in all cases. Among them, VGG+SpecFormer achieved the best results, which may be because VGG can discriminate the widest range of frequency masking, while SpecFormer compensates for its inability to discriminate high-frequency noise (please refer to the supplementary for details).

\section{Discussion and Limitations}
With the introduction of an additional spectral discriminator, our obtained SR images have their spectra better aligned to those of the real images, thereby enhancing the overall perceptual quality of our methodology. Despite these benefits, we notice that better-aligned spectra may not always result in better perceptual quality. Further, the current method trains the spatial and spectral discriminator separately, increasing the computational and storage overhead. These issues indicate that there is still much room for improvement of the spectral discriminators.

\section{Conclusion}
This paper investigates the effectiveness of spectral discriminators. Our research reveals that spatial and spectral discriminators offer unique benefits, and they work best when being used together. We also introduce a per-patch Fourier Transform to improve the spectral discriminator's structure. Our extensive experiments on image SR and NR-IQA confirm the effectiveness of our approach. Specifically, our method improves the alignment of spectra and the PD tradeoff in SR, as well as the IQA ability in NR-IQA.

{\small
	\bibliographystyle{ieee_fullname}
	\bibliography{egbib}
}
\clearpage

\appendix

\section{Efficiency Analysis of Discriminators}
To compare the efficiency differences between the Spectral Transformer and other commonly used discriminators in SR, we consider three metrics: the total number of parameters, the number of floating point operations~(FLOPs), and the number of elements of all outputs of convolutional layers~(activations). As is evidenced in \cref{tab:tab8}, our discriminator is, in fact, highly efficient due to our utilization of a small number of dimensions and a relatively large patch size. For our SR experiments, we employed a patch size of $32\times 32$ for both the Spatial Transformer and the Spectral Transformer. Therefore, the number of parameters in our discriminator is 4.4M, the number of FLOPs is 186.82G, and the number of activations is 1.26G. Among these, the number of parameters in our discriminator is equivalent to that of U-Net, while the FLOPs and Activations are significantly lower than those of VGG and U-Net.

\begin{table}[t]
	\setlength{\abovecaptionskip}{3pt}
	\centering
	\resizebox{\linewidth}{12.5mm}{
		\begin{tabular}{cccccc}
			\toprule
			Discriminator & Params[M] & FLOPs[G] & Activations[G] \\
			\midrule
			VGG~\cite{wang2018esrgan} & 21.1 & 8728.63 & 9.57  \\
			U-Net~\cite{wang2021real} & 4.4 & 24776.00 & 22.56  \\
			SpecFormer/8 & 2.0 & 2709.60 & 33.32  \\
			SpecFormer/32 & 2.2 & 93.41 & 0.63 \\
			\bottomrule
		\end{tabular}
	}
	\caption{\textbf{Efficiency performance of various discriminators.} Metrics are evaluated on images of size $256\times 256$.}
	\label{tab:tab8}
\end{table}

\begin{table}[t]
	\centering
	\resizebox{\linewidth}{7.5mm}{
		\begin{tabular}{cccc}
			\toprule
			& Ground Truth & Low-Quality & Super-Resolution \\
			\midrule
			Real-ESRGAN~\cite{wang2021real} & 0.77 & 0.92 & 0.16 \\
			Real-ESRGAN + SpecFormer & 0.52 & 0.43 & 0.44 \\
			\bottomrule
		\end{tabular}
	}
	\caption{\textbf{The average scores of discriminators} \wrt three types of images. The spectral discriminator, \ie, SpecFormer, mitigates the high-frequency flaw of Real-ESRGAN's spatial discriminator.}
	\label{tab:tab9}
	\vspace{-1em}
\end{table}

\section{The Spectral Discriminator solve the high-frequency noise problem}
It is the extreme preference of Real-ESRGAN's spatial discriminator for high-frequency information that motivated us to introduce the spectral discriminator. To confirm that the spectral discriminator can address this problem in practical applications, we introduce spectral Transformer to train Real-ESRGAN. As demonstrated in \cref{tab:tab9}, our approach yields the highest scores for ground truth images, followed by super-resolution, and the lowest scores for low-quality images, as we had anticipated. Therefore, we can conclude that the spectral discriminator is capable of mitigating the flaw of the spatial discriminator on high frequencies.

\section{Spectral Profile of SR Models under Bicubic Degradation}
\begin{figure}[t]
	\centering
	\includegraphics[width=\linewidth]{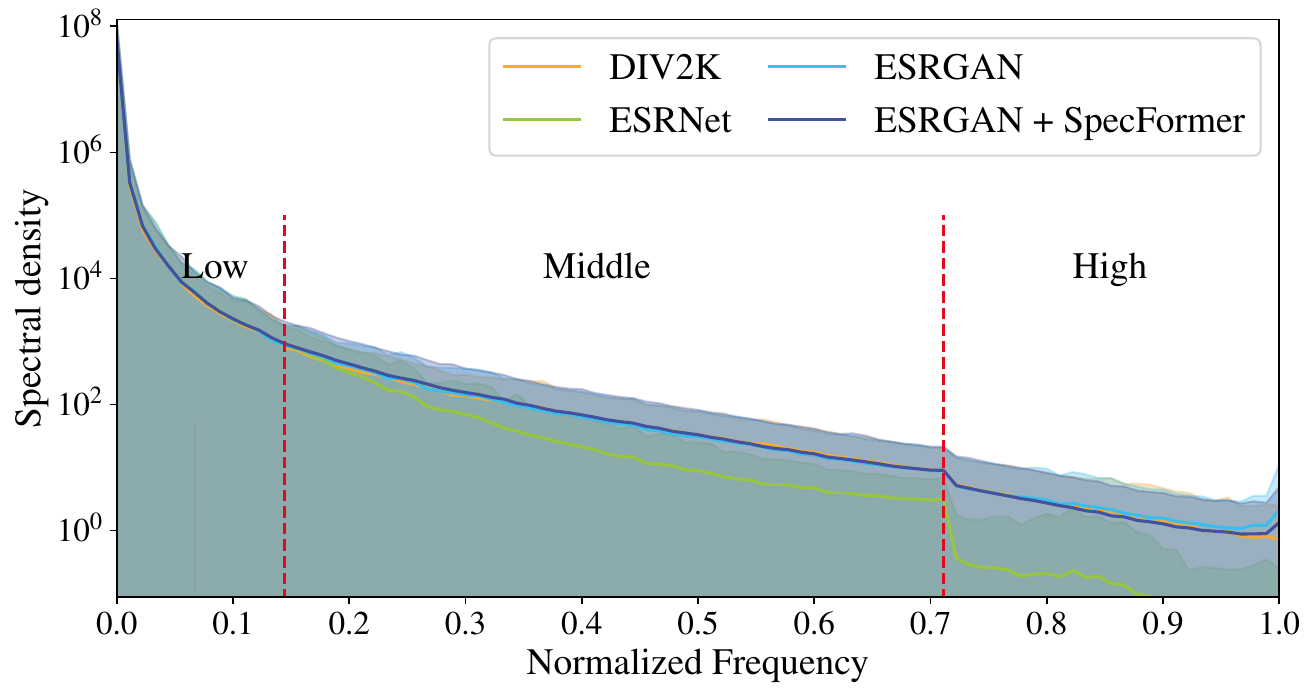}
	\caption{\textbf{Spectral profile of SR models under bicubic degradation}. The spatial discriminator could match the real spectra well under simple degradation~(bicubic downsampling).}
	\label{fig:fig8}
	\vspace{-1em}
\end{figure}
We have observed that SR networks exhibit poor spectral alignment with real spectra in real-world SR scenarios, which prompted us to introduce a spectral discriminator to improve the spectral alignment of SR networks. Nevertheless, we acknowledge that this issue is not as severe in the case of simple degradation, such as bicubic degradation. As depicted in \cref{fig:fig8}, the SR images produced by ESRGAN~\cite{wang2018esrgan} already match real images in the low and middle-frequency ranges. While our Spectral Transformer mitigates the problem of excessive preference for high-frequency content by the spatial discriminator to some extent, its impact on quantitative metrics and human perception may not be significant.

\begin{figure}[t]
	\centering
	\begin{subfigure}[b]{0.32\linewidth}
		\centering
		\includegraphics[width=\linewidth]{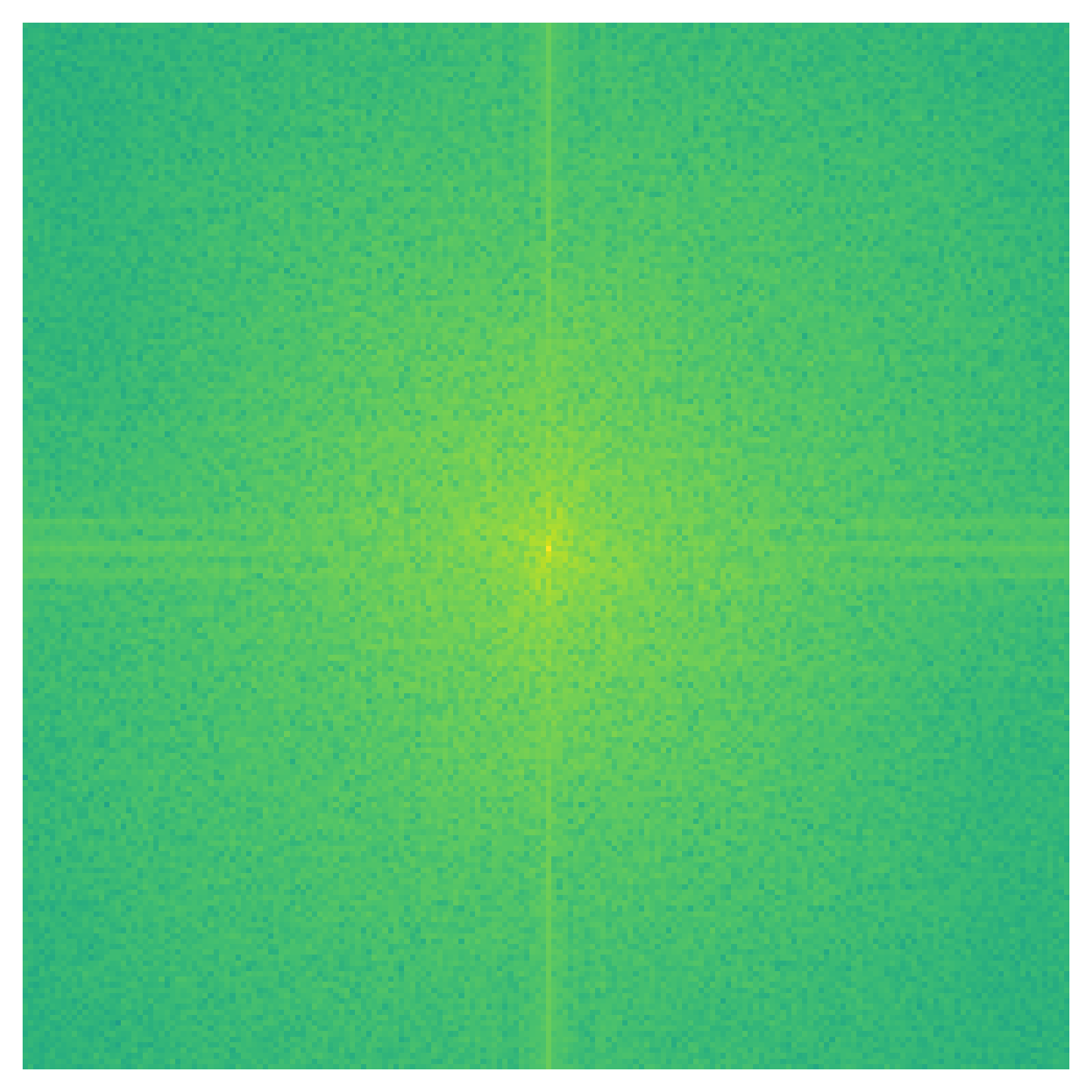}
		\caption{GT}
		\label{fig:fig9:a}
	\end{subfigure}
	\begin{subfigure}[b]{0.32\linewidth}
		\centering
		\includegraphics[width=\linewidth]{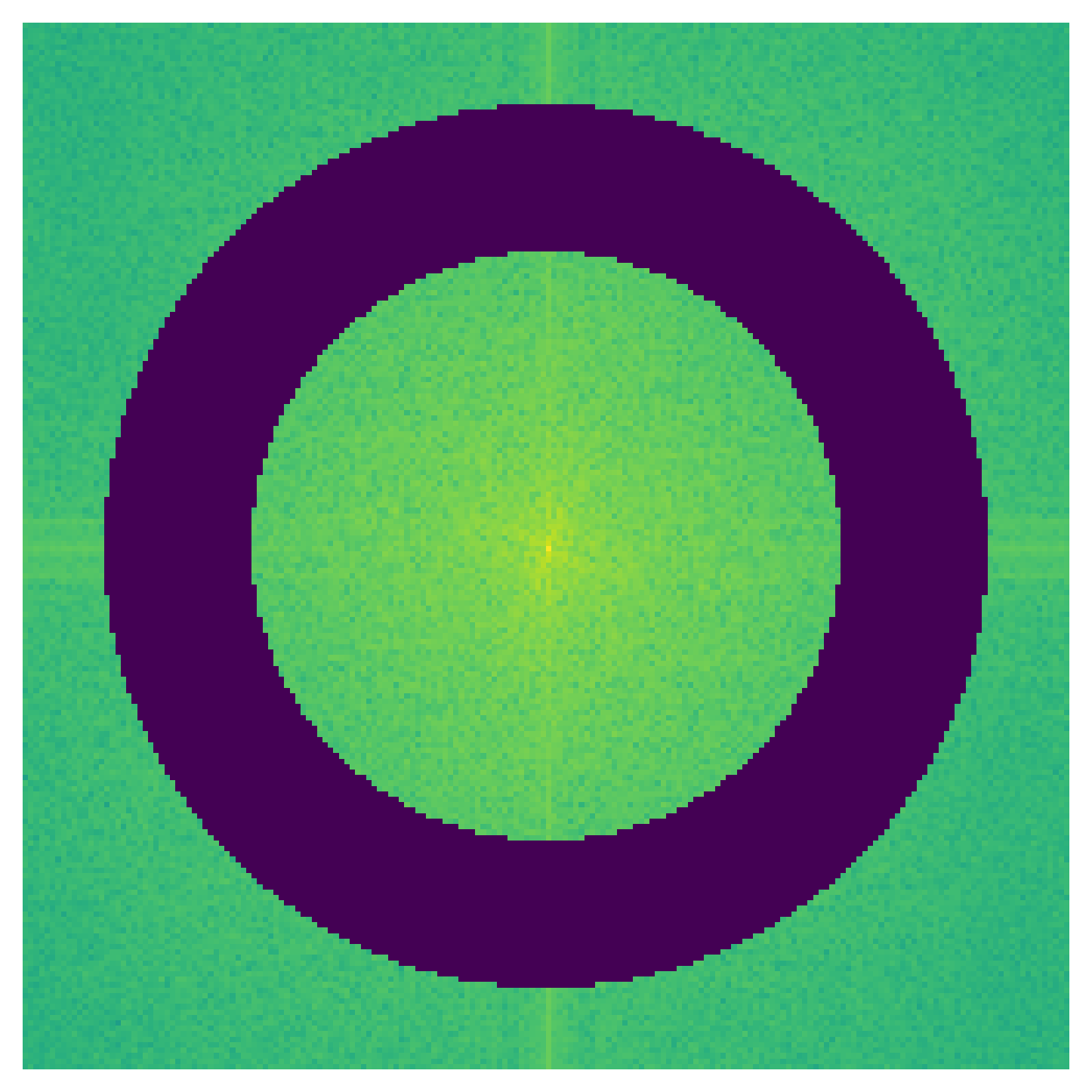}
		\caption{After masking}
		\label{fig:fig9:b}
	\end{subfigure}
	\begin{subfigure}[b]{0.32\linewidth}
		\centering
		\includegraphics[width=\linewidth]{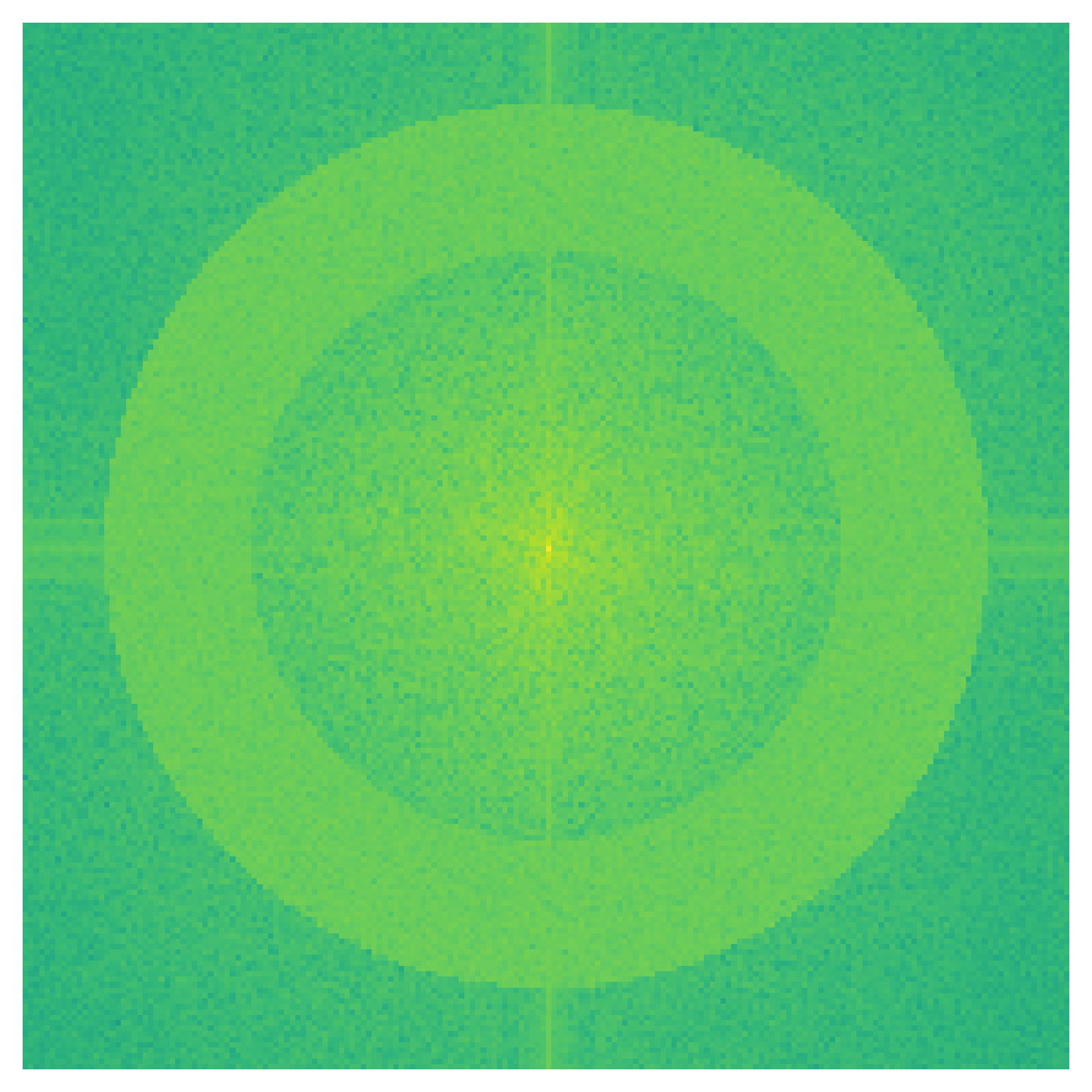}
		\caption{After noise}
		\label{fig:fig9:c}
	\end{subfigure}
	\caption{\textbf{Intuitive visualization of frequency masking (b) and noise (c).}}
	\vspace{-0.5em}
	\label{fig:fig9}
\end{figure}

\section{The Visual Effects of Frequency Perturbations}
We investigated the behavior of discriminator by examining its performance under two representative frequency perturbations, and found differences in the capabilities of spatial and spectral discriminators. Here, we provide more comprehensible visualizations, as exemplified in \cref{fig:fig9}, where frequency masking is the removal of a circular ring with a certain radius from the spectrogram, and frequency noise is the addition of noise within a circular ring with a certain radius in the spectrogram. In addition, we also demonstrate the effect of frequency perturbation on two representative images in \cref{fig:fig13}. Among them, high-frequency perturbations are relatively difficult for the human eye to perceive, while other perturbations have a significant impact on human perception.

\begin{figure*}[t]
	\centering
	\includegraphics[width=\linewidth]{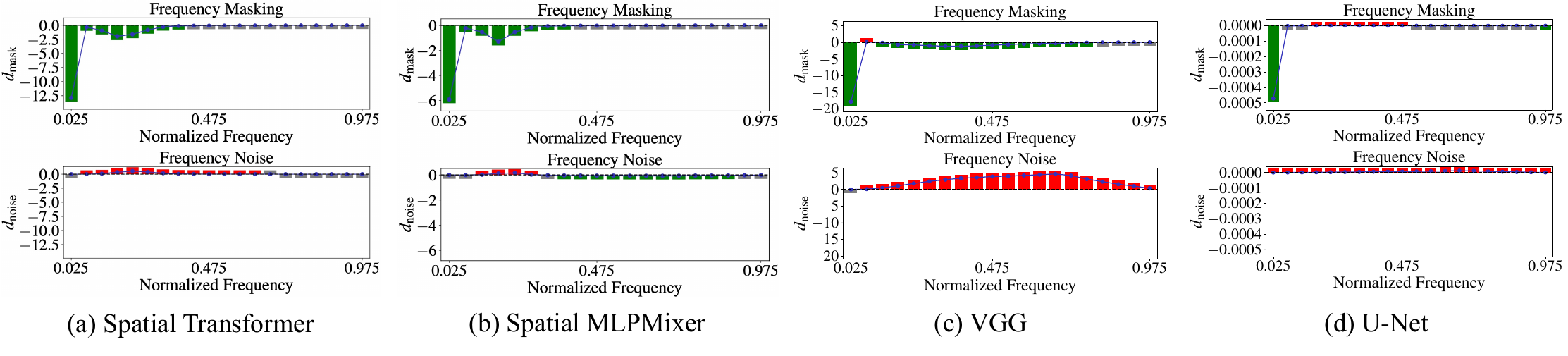}
	\caption{\textbf{Robustness behavior of various spatial discriminators}. The Transformer/MLP-Mixer works better at identifying absence in the middle-frequency range, and the CNN is aware of the higher-range spectrum. Also, CNN works poorly at identifying frequency noise compared to Transformer/MLP-Mixer. The U-Net has lower spectra perception compared to VGG due to its residual structure~\cite{wang2020high}.}
	\vspace{-0.5em}
	\label{fig:fig10}
\end{figure*}

\begin{figure*}[t]
	\centering
	\begin{subfigure}[b]{0.33\linewidth}
		\centering
		\includegraphics[width=\linewidth]{figures/test_robustness_on_frequency_20_spectral_transformer.pdf}
		\caption{Spectral Transformer}
		\label{fig:fig11:a}
	\end{subfigure}
	\begin{subfigure}[b]{0.33\linewidth}
		\centering
		\includegraphics[width=\linewidth]{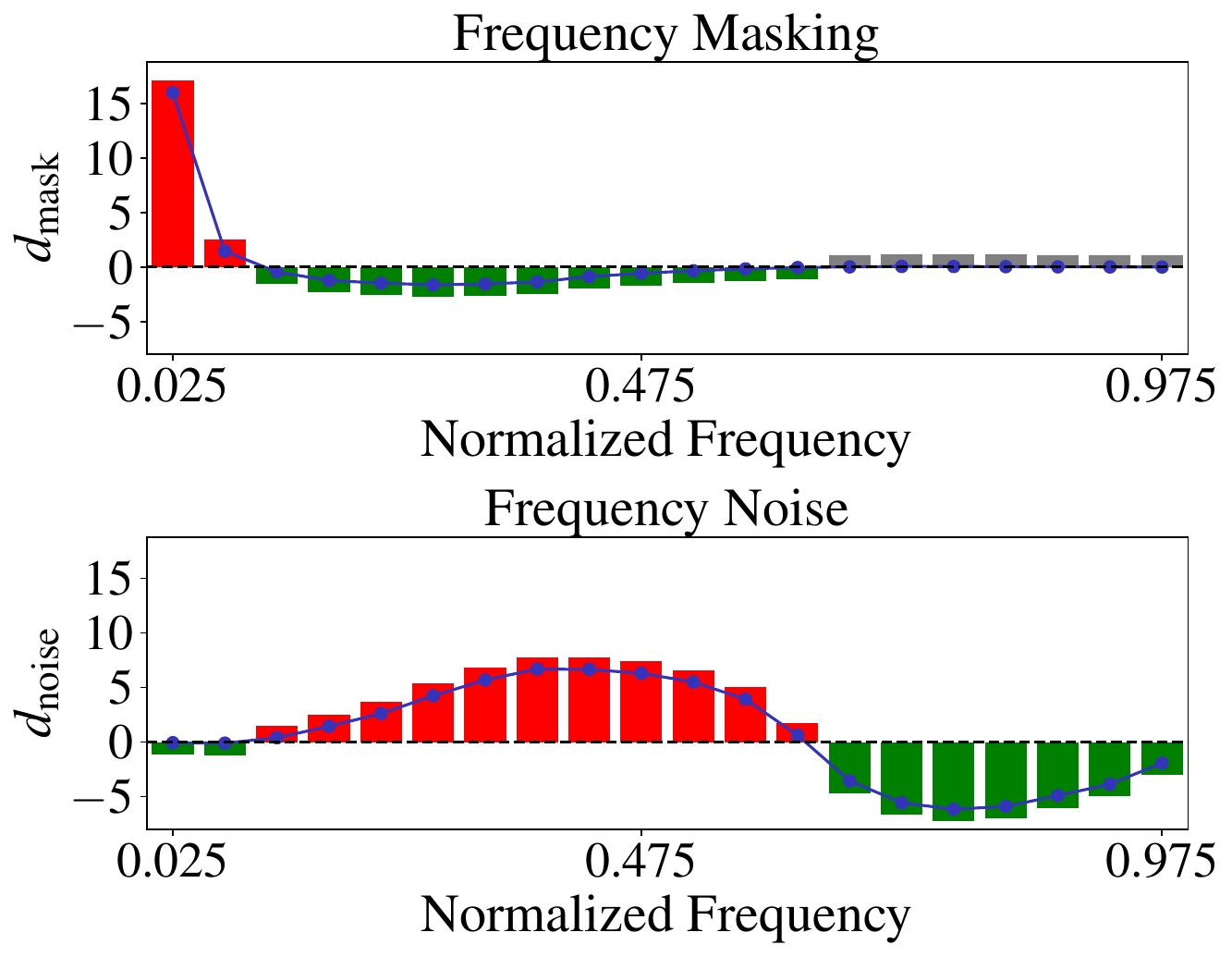}
		\caption{Spectral MLP-Mixer}
		\label{fig:fig11:b}
	\end{subfigure}
	\begin{subfigure}[b]{0.33\linewidth}
		\centering
		\includegraphics[width=\linewidth]{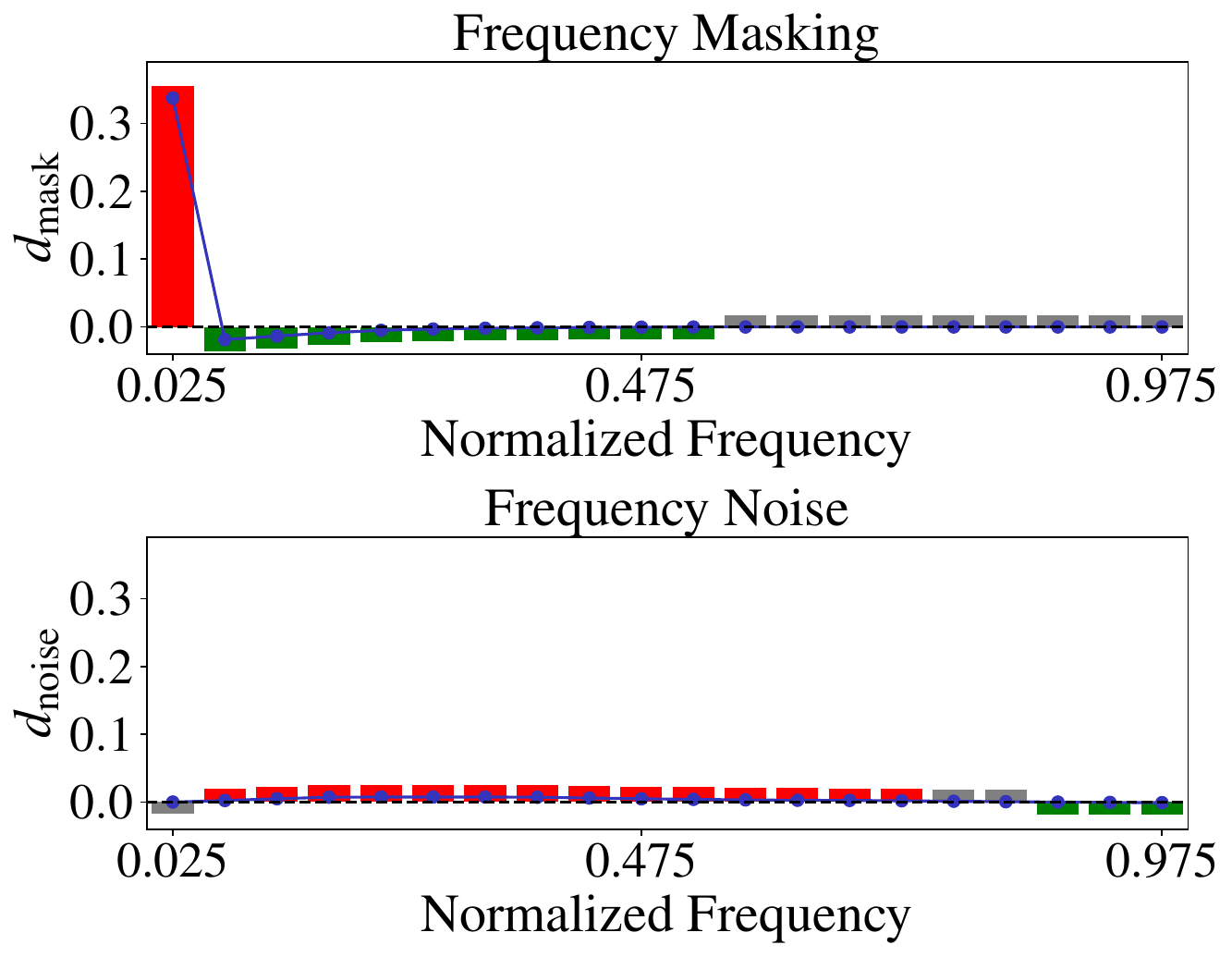}
		\caption{Spectral MLP}
		\label{fig:fig11:c}
	\end{subfigure}
	\caption{\textbf{Robustness behavior of various spectral discriminators}. Overall, these architectures exhibit similar three-stage behaviors. Specifically, Transformer and MLP-Mixer perform almost identically, while MLP fails to learn to discriminate high-frequency noise effectively.}
	\vspace{-0.5em}
	\label{fig:fig11}
\end{figure*}

\section{Architecture-Related Robustness}

In main body of the text, we argue that the spatial discriminator excels at identifying low-frequency masking, while the spectral discriminator is good at identifying high-frequency noise. This section substantiates that the aforementioned phenomenon is independent of the specific network architecture. 

\subsection{Spatial Discriminators}
\cref{fig:fig10} illustrates the robustness of various spatial discriminators under frequency perturbations. All four representative network architectures demonstrate a similar tendency to capture low-frequency masking. Nevertheless, subtle yet critical distinctions exist between these architectures. While Transformer performs like a low-pass filter~\cite{park2022how}, relying more on low-frequency information, it can identify a narrower range of frequency masking than typical CNNs such as VGG~\cite{simonyan2014very}. Similarly, MLP-Mixer behaves like Transformer due to their similar high-level architecture design. In contrast, VGG, which is a CNN network, has a broader spectrum perception range. The U-Net~\cite{ronneberger2015u}, which is a residual structured network, has a weaker tendency to capture high-frequency components~\cite{wang2020high}, and therefore behaves more like the Transformer~\cite{dosovitskiy2021an}/MLP-Mixer~\cite{tolstikhin2021mlp}. These phenomena align with those observed in other studies~\cite{wang2020high,naseer2021intriguing,shao2021adversarial,benz2021adversarial}.

\begin{figure*}[t]
	\centering
	\begin{subfigure}[b]{0.245\linewidth}
		\centering
		\includegraphics[width=\linewidth]{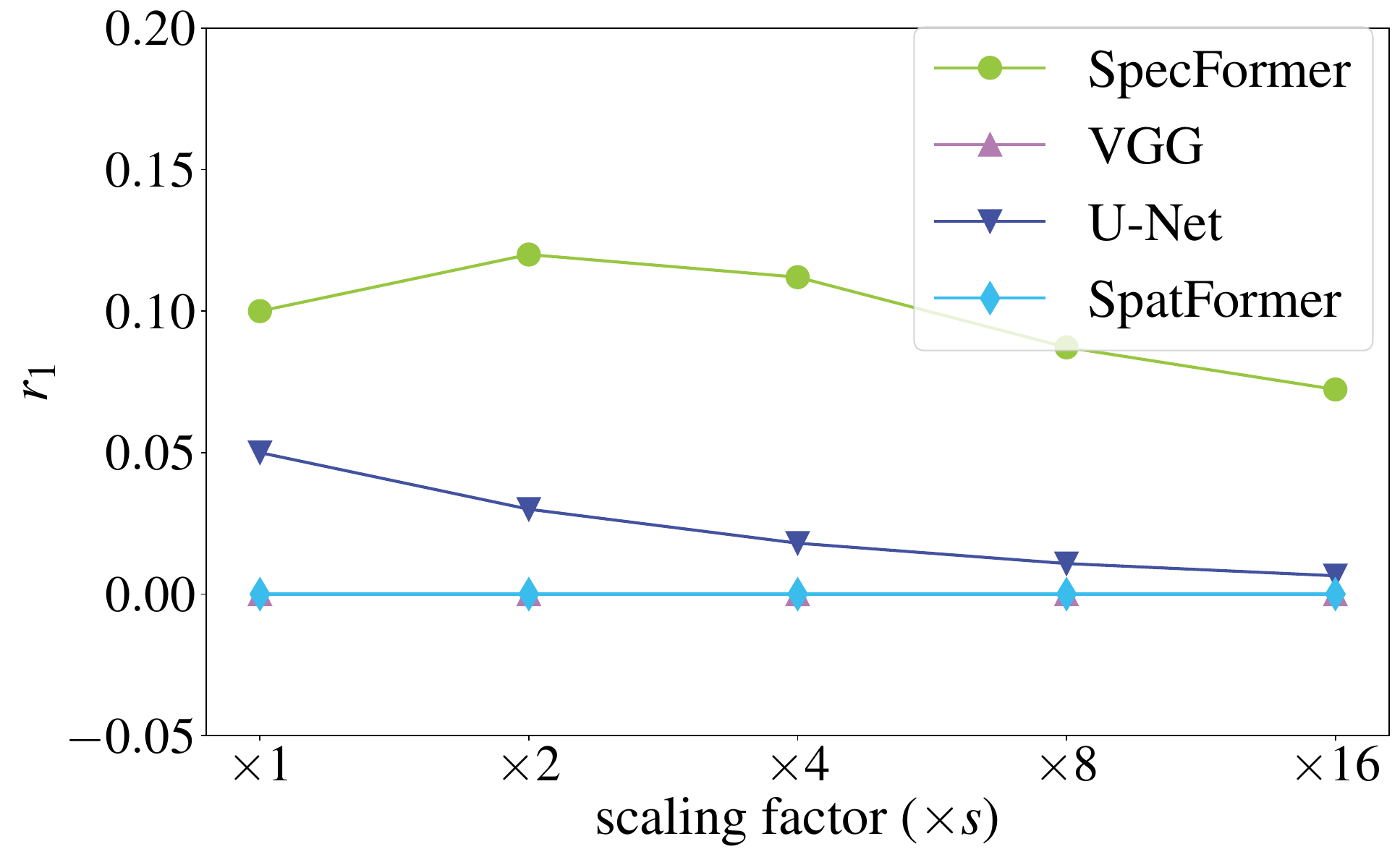}
		\caption{$r_1$ on frequency masking}
		\label{fig:fig12:a}
	\end{subfigure}
	\begin{subfigure}[b]{0.245\linewidth}
		\centering
		\includegraphics[width=\linewidth]{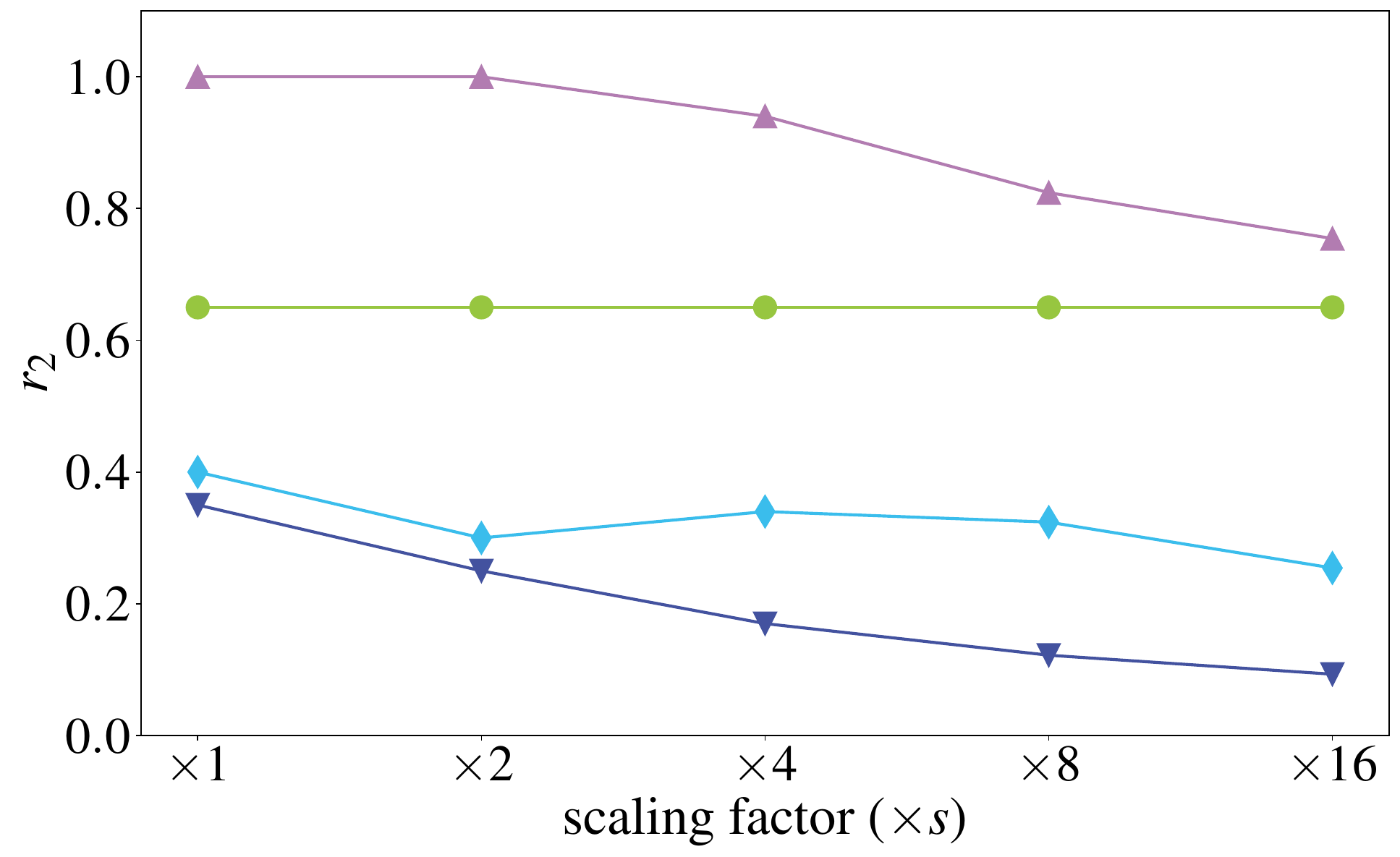}
		\caption{$r_2$ on frequency masking}
		\label{fig:fig12:b}
	\end{subfigure}
	\begin{subfigure}[b]{0.245\linewidth}
		\centering
		\includegraphics[width=\linewidth]{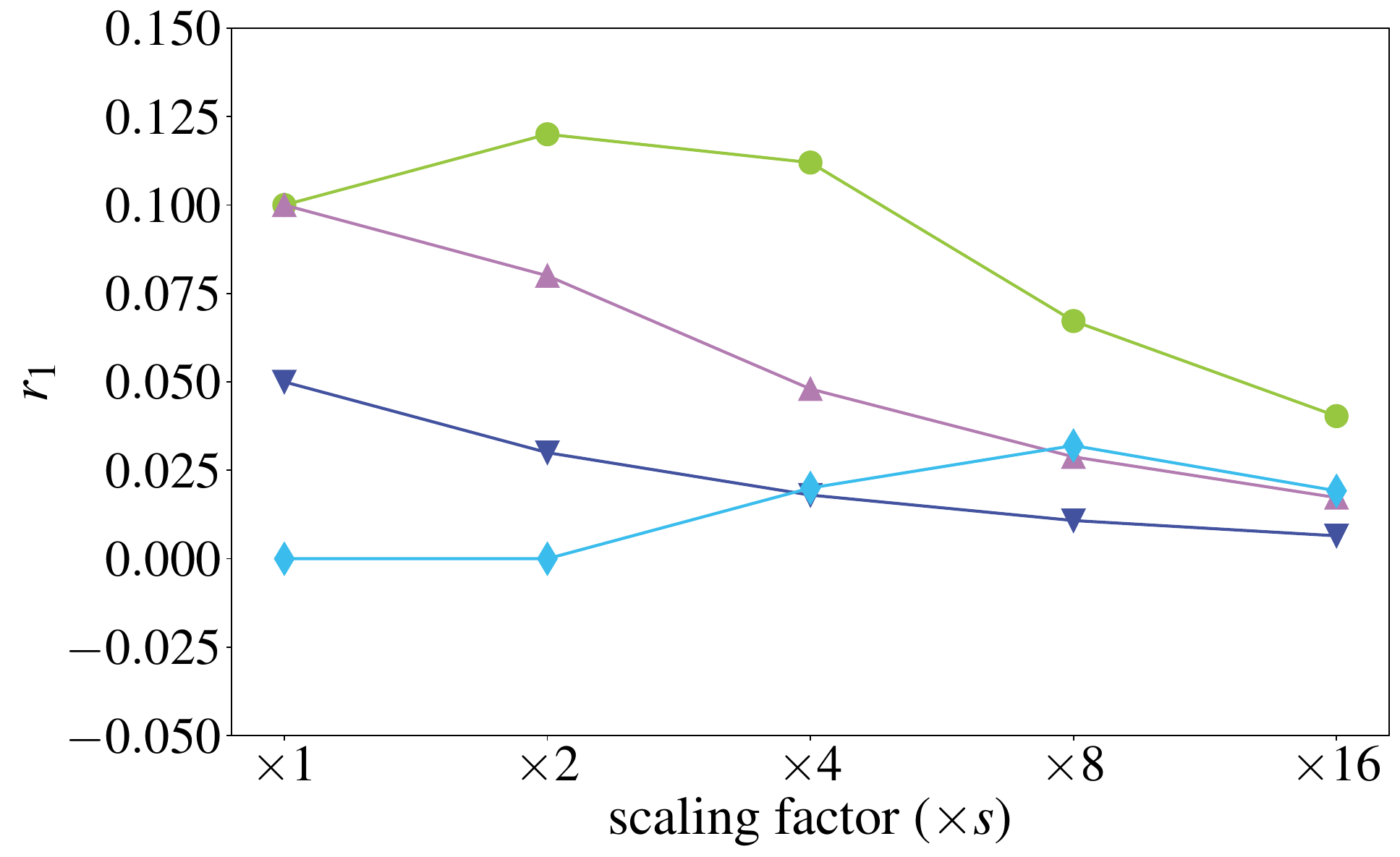}
		\caption{$r_1$ on frequency noise}
		\label{fig:fig12:c}
	\end{subfigure}
	\begin{subfigure}[b]{0.245\linewidth}
		\centering
		\includegraphics[width=\linewidth]{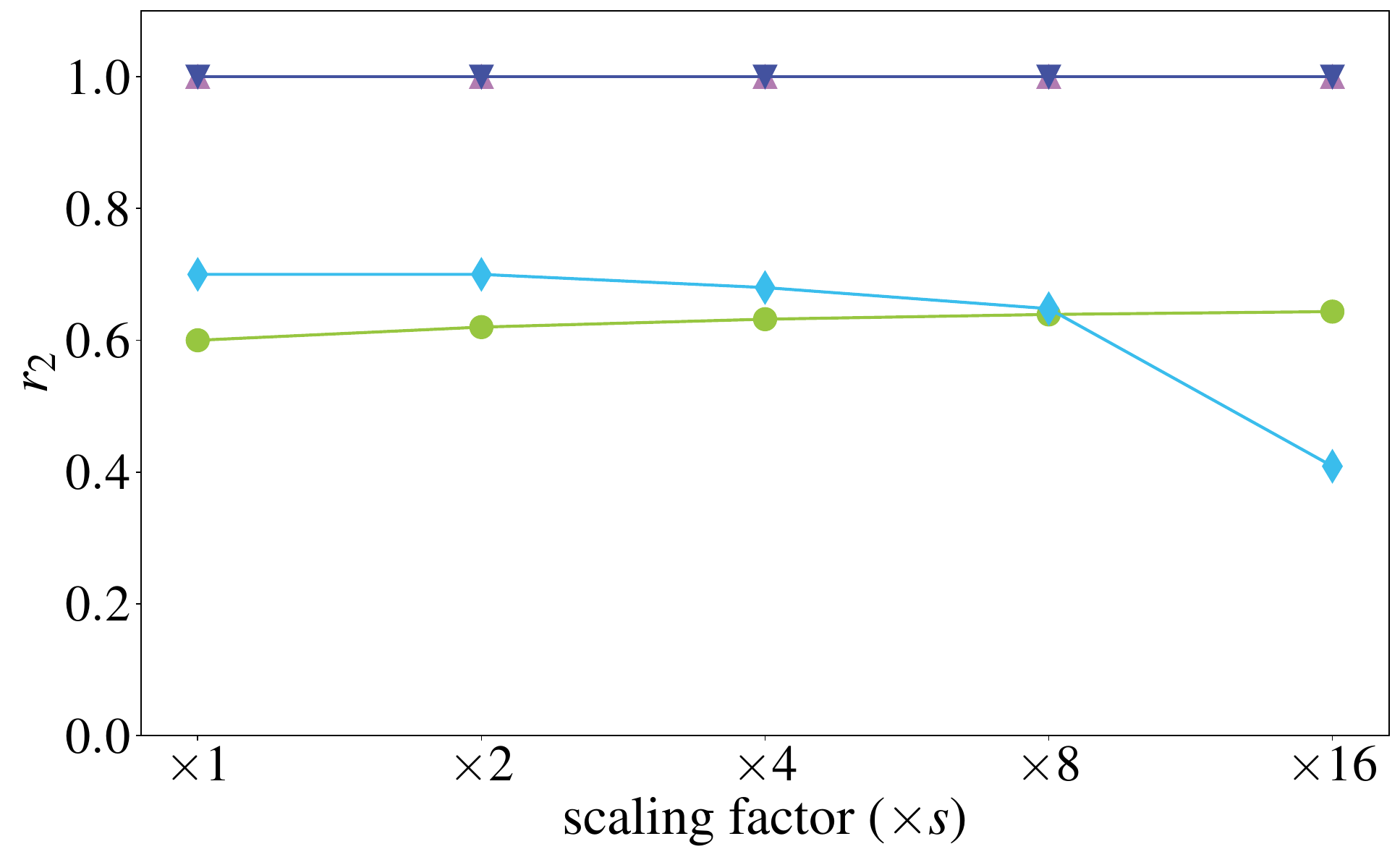}
		\caption{$r_2$ on frequency noise}
		\label{fig:fig12:d}
	\end{subfigure}
	
	\caption{\textbf{The shifting behavior of the three frequency ranges varying scaling factors}. The three frequency ranges are $\left[0, r_1\right)$, $\left[r_1, r_2\right)$, and $\left[r_2, 1\right]$. SpatFormer/SpecFormer denotes Transformer applied to the spatial/frequency domain. As the scaling factor grows, the optimization goal of generator migrates from distortion to perception (similar to increasing the weight of the perception term). Consequently, the boundaries of the three frequency ranges shift to the left ($r_{1}$ and $r_{2}$ decrease). The tiny vibrations may be related to the stochasticity of the training.}
	\label{fig:fig12}
\end{figure*}

\subsection{Spectral Discriminators}
As evidenced by \cref{fig:fig11}, comparable to the scenario of spatial discriminators, spectral discriminators also exhibit similar behaviors,~\ie, they are unable to differentiate the absence of low frequencies. Specifically, just as they do in the spatial domain, both Transformer and MLP-Mixer exhibit consistent behavior in the frequency domain, as they both effectively learn to discriminate against high-frequency noise. While Spectral MLP performs similarly to Transformer/MLP-Mixer in terms of frequency masking, it fails to learn to recognize high-frequency noise, further validating the effectiveness of our Spectral Transformer.

In conclusion, there exists a fundamental disparity between the spatial and spectral discriminator. Specifically, the spatial discriminator is an expert at discriminating low-frequency masking, while the spectral discriminator performs better in distinguishing high-frequency noise, and the architecture contributes to specific behavior. Therefore, it is crucial to consider the specific requirements of the task and the characteristics of the input data when choosing a discriminator architecture. Moreover, our findings can guide future research in developing discriminators that are better suited for specific tasks and data types.

\section{The Generalizability of the Three Frequency Ranges Phenomenon}
We have demonstrated that both the generator and discriminator exhibit three-range behavior in the frequency domain, and we have explained this phenomenon from the frequency perspective of the PD tradeoff. Nevertheless, in the main body of the text, we conducted our study using a $\times 4$ SR as an example. In order to demonstrate the generalizability of the three-range behavior in the frequency domain, we investigated how the scaling factor of SR influences various discriminators. Specifically, we defined the boundary of the three frequency ranges as $r_1$ and $r_2$, where $r_{1}\geq 0$, $r_1\leq r_{2}$, and $r_{2}\leq 1$. These three frequency ranges are in the radius intervals $\left[0,r_{1}\right)$, $\left[r_{1},r_{2}\right)$, and $\left[r_{2},1\right]$, respectively. Please refer to \cref{fig:fig11:a} for an illustration of the properties of each range.

Let's start by taking a global view. \cref{fig:fig12} shows the boundary of the three frequency ranges ($r_1$ and $r_2$), which will shift to the left as the scaling factor $s$ increases. This can be explained from the frequency perspective of PD tradeoff. Initially, as the scaling factor $s$ increases, the information accessible in the input image diminishes. Subsequently, the low-frequency part that the generator can perfectly recover also decreases, and the perception term gradually dominates optimization. As a result, $r_{1}$ decreases. Moreover, the limited capacity of the generator can cause a decrease in $r_{2}$, considering the decrease of $r_{1}$, though the decreasing trend of $r_{2}$ is relatively mild compared to $r_{1}$. Furthermore, when $s$ approaches infinity ($\times\infty$ SR), the input contains scarcely any information, thereby equivalent to an unconditional image generation task. In this scenario, the discriminator can identify a significant amount of frequency noise but can only discriminate a small fraction of the frequency masking. Chen~\etal~\cite{chen2021ssd} also observed this limiting case in image generation.

\begin{figure*}[t]
	\centering
	\begin{subfigure}[b]{0.16\linewidth}
		\centering
		\includegraphics[width=\linewidth]{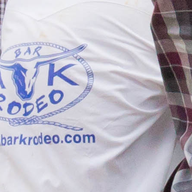}
		\caption{GT}
		\label{fig:fig13:a}
	\end{subfigure}
	\begin{subfigure}[b]{0.16\linewidth}
		\centering
		\includegraphics[width=\linewidth]{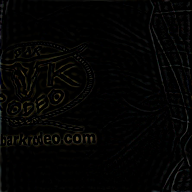}
		\caption{masking $\left[0, \frac{1}{5}\right]$}
		\label{fig:fig13:b}
	\end{subfigure}
	\begin{subfigure}[b]{0.16\linewidth}
		\centering
		\includegraphics[width=\linewidth]{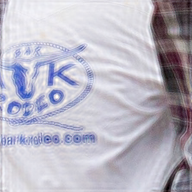}
		\caption{masking $\left[\frac{1}{5}, \frac{2}{5}\right]$}
		\label{fig:fig13:c}
	\end{subfigure}
	\begin{subfigure}[b]{0.16\linewidth}
		\centering
		\includegraphics[width=\linewidth]{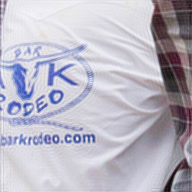}
		\caption{masking $\left[\frac{2}{5}, \frac{3}{5}\right]$}
		\label{fig:fig13:d}
	\end{subfigure}
	\begin{subfigure}[b]{0.16\linewidth}
		\centering
		\includegraphics[width=\linewidth]{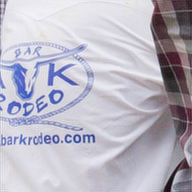}
		\caption{masking $\left[\frac{3}{5}, \frac{4}{5}\right]$}
		\label{fig:fig13:e}
	\end{subfigure}
	\begin{subfigure}[b]{0.16\linewidth}
		\centering
		\includegraphics[width=\linewidth]{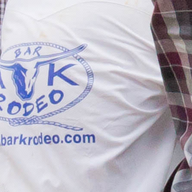}
		\caption{masking $\left[\frac{4}{5}, 1\right]$}
		\label{fig:fig13:f}
	\end{subfigure}
	\begin{subfigure}[b]{0.16\linewidth}
		\centering
		\includegraphics[width=\linewidth]{figures/3.png}
		\caption{GT}
		\label{fig:fig13:g}
	\end{subfigure}
	\begin{subfigure}[b]{0.16\linewidth}
		\centering
		\includegraphics[width=\linewidth]{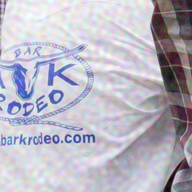}
		\caption{noise $\left[0, \frac{1}{5}\right]$}
		\label{fig:fig13:h}
	\end{subfigure}
	\begin{subfigure}[b]{0.16\linewidth}
		\centering
		\includegraphics[width=\linewidth]{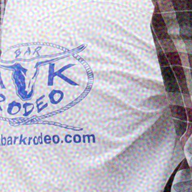}
		\caption{noise $\left[\frac{1}{5}, \frac{2}{5}\right]$}
		\label{fig:fig13:i}
	\end{subfigure}
	\begin{subfigure}[b]{0.16\linewidth}
		\centering
		\includegraphics[width=\linewidth]{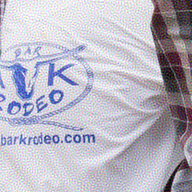}
		\caption{noise $\left[\frac{2}{5}, \frac{3}{5}\right]$}
		\label{fig:fig13:j}
	\end{subfigure}
	\begin{subfigure}[b]{0.16\linewidth}
		\centering
		\includegraphics[width=\linewidth]{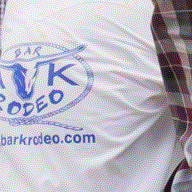}
		\caption{noise $\left[\frac{3}{5}, \frac{4}{5}\right]$}
		\label{fig:fig13:k}
	\end{subfigure}
	\begin{subfigure}[b]{0.16\linewidth}
		\centering
		\includegraphics[width=\linewidth]{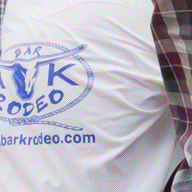}
		\caption{noise $\left[\frac{4}{5}, 1\right]$}
		\label{fig:fig13:l}
	\end{subfigure}
	\begin{subfigure}[b]{0.16\linewidth}
		\centering
		\includegraphics[width=\linewidth]{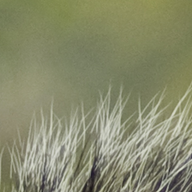}
		\caption{GT}
		\label{fig:fig13:m}
	\end{subfigure}
	\begin{subfigure}[b]{0.16\linewidth}
		\centering
		\includegraphics[width=\linewidth]{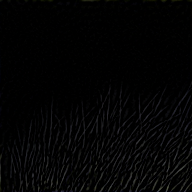}
		\caption{masking $\left[0, \frac{1}{5}\right]$}
		\label{fig:fig13:n}
	\end{subfigure}
	\begin{subfigure}[b]{0.16\linewidth}
		\centering
		\includegraphics[width=\linewidth]{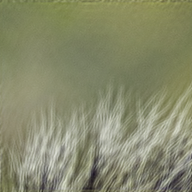}
		\caption{masking $\left[\frac{1}{5}, \frac{2}{5}\right]$}
		\label{fig:fig13:o}
	\end{subfigure}
	\begin{subfigure}[b]{0.16\linewidth}
		\centering
		\includegraphics[width=\linewidth]{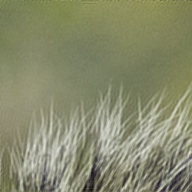}
		\caption{masking $\left[\frac{2}{5}, \frac{3}{5}\right]$}
		\label{fig:fig13:p}
	\end{subfigure}
	\begin{subfigure}[b]{0.16\linewidth}
		\centering
		\includegraphics[width=\linewidth]{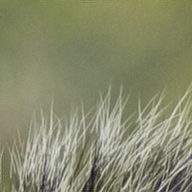}
		\caption{masking $\left[\frac{3}{5}, \frac{4}{5}\right]$}
		\label{fig:fig13:q}
	\end{subfigure}
	\begin{subfigure}[b]{0.16\linewidth}
		\centering
		\includegraphics[width=\linewidth]{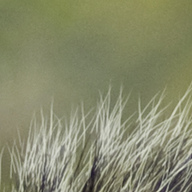}
		\caption{masking $\left[\frac{4}{5}, 1\right]$}
		\label{fig:fig13:r}
	\end{subfigure}
	\begin{subfigure}[b]{0.16\linewidth}
		\centering
		\includegraphics[width=\linewidth]{figures/9.png}
		\caption{GT}
		\label{fig:fig13:s}
	\end{subfigure}
	\begin{subfigure}[b]{0.16\linewidth}
		\centering
		\includegraphics[width=\linewidth]{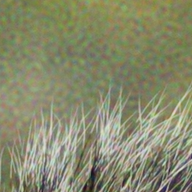}
		\caption{noise $\left[0, \frac{1}{5}\right]$}
		\label{fig:fig13:t}
	\end{subfigure}
	\begin{subfigure}[b]{0.16\linewidth}
		\centering
		\includegraphics[width=\linewidth]{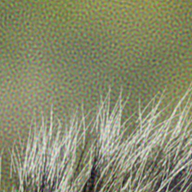}
		\caption{noise $\left[\frac{1}{5}, \frac{2}{5}\right]$}
		\label{fig:fig13:u}
	\end{subfigure}
	\begin{subfigure}[b]{0.16\linewidth}
		\centering
		\includegraphics[width=\linewidth]{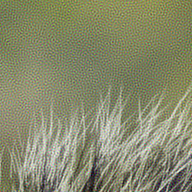}
		\caption{noise $\left[\frac{2}{5}, \frac{3}{5}\right]$}
		\label{fig:fig13:v}
	\end{subfigure}
	\begin{subfigure}[b]{0.16\linewidth}
		\centering
		\includegraphics[width=\linewidth]{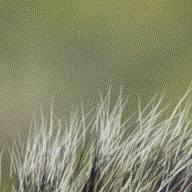}
		\caption{noise $\left[\frac{3}{5}, \frac{4}{5}\right]$}
		\label{fig:fig13:w}
	\end{subfigure}
	\begin{subfigure}[b]{0.16\linewidth}
		\centering
		\includegraphics[width=\linewidth]{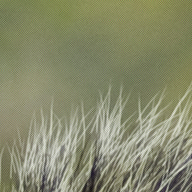}
		\caption{noise $\left[\frac{4}{5}, 1\right]$}
		\label{fig:fig13:x}
	\end{subfigure}
	\caption{\textbf{The effects of frequency masking and noise on two representative images}.}
	\label{fig:fig13}
\end{figure*}

\end{document}